\documentclass[journal]{IEEEtran}
\ifCLASSINFOpdf
\else
\fi
\usepackage{amsmath, amsthm, amssymb}
\usepackage{algorithm}
\usepackage{multirow}
\usepackage{array}
\usepackage{booktabs}
\usepackage{bm}
\usepackage{graphicx}
\usepackage{xcolor}
\usepackage{soul}
\usepackage{hyperref}
\usepackage{comment}
\usepackage{fixmath}

\usepackage{subcaption}
\usepackage{colortbl}
\usepackage{cases}
\usepackage[normalem]{ulem}
\begin{document}
%
\title{Operational Support Estimator Networks}
%
%
%

\author{Mete Ahishali, Mehmet Yamac, Serkan Kiranyaz, Moncef Gabbouj
\thanks{Mete Ahishali, Mehmet Yamac, and Moncef Gabbouj are with the Faculty of Information Technology and Communication Sciences, Tampere University, Tampere, Finland (email: \textit{name.surname@tuni.fi}).}
\thanks{Serkan Kiranyaz is with the Department of Electrical Engineering, Qatar University, Doha, Qatar (email: \textit{mkiranyaz@qu.edu.qa}).}
}

\maketitle

\begin{abstract}
In this work, we propose a novel approach called Operational Support Estimator Networks (OSENs) for the support estimation task. Support Estimation (SE) is defined as finding the locations of non-zero elements in sparse signals. By its very nature, the mapping between the measurement and sparse signal is a non-linear operation. Traditional support estimators rely on computationally expensive iterative signal recovery techniques to achieve such non-linearity. Contrary to the convolutional layers, the proposed OSEN approach consists of operational layers that can learn such complex non-linearities without the need for deep networks. In this way, the performance of non-iterative support estimation is greatly improved. Moreover, the operational layers comprise so-called generative \textit{super neurons} with non-local kernels. The kernel location for each neuron/feature map is optimized jointly for the SE task during training. We evaluate the OSENs in three different applications: i. support estimation from Compressive Sensing (CS) measurements, ii. representation-based classification, and iii. learning-aided CS reconstruction where the output of OSENs is used as prior knowledge to the CS algorithm for enhanced reconstruction. Experimental results show that the proposed approach achieves computational efficiency and outperforms competing methods, especially at low measurement rates by significant margins. The software implementation is shared at \href{https://github.com/meteahishali/OSEN}{https://github.com/meteahishali/OSEN}.
\end{abstract}


%
\IEEEpeerreviewmaketitle

\section{Introduction}

\label{sec:introduction}

\IEEEPARstart{S}{parse} Representation of a signal $\mathbf{y} \in \mathbb{R}^{m}$ defines that the signal is represented as a linear combination of only a small subset of $k$ atoms from the entire dictionary with $n$ elements, where $k$ is significantly smaller than the total number of atoms, i.e.,  $k<<n$. Mathematically speaking, let $\mathbf{D} \in \mathbb{R}^{m \times n}$ is an underdetermined matrix where $m < n$ and the following representation: $\mathbf{y} = \mathbf{Dx}$ is considered sparse if there are a few non-zero coefficients in the representation vector $\mathbf{x}$. Finding indices of these non-zero coefficients forms the task of (Sparse) Support Estimation (SE) \cite{SE1,SE3}. Alternatively, SE can be defined as the localization of the subset consisting of smallest number of bases atoms whose linear combination with the corresponding representation coefficients forms the original signal. There is a fundamental difference between SE and traditional (Sparse) Signal Recovery/Reconstruction (SR). In the latter, the aim is to find the exact values of $\mathbf{x}$. Hence, it is a more challenging task than the SE as it also includes the estimation of the representation coefficients in addition to the localization.

In general, many tasks can be formulated as an SR task, for example, consider Compressive Sensing (CS) problem \cite{CS1,CS2} $\mathbf{y} = \mathbf{As}$ for a signal $\mathbf{s} \in \mathbb{R}^d$ where $\mathbf{A} \in \mathbb{R}^{m \times d}$ is the measurement matrix. If the signal $\mathbf{s}$ is sparsely coded in a proper domain $\mathbf{\Phi} \in \mathbb{R}^{d \times n}$ i.e, $\mathbf{s} = \mathbf{\Phi} \mathbf{x}$, then linear measurement system can be expressed as $\mathbf{y} = \mathbf{Dx}$ where $\mathbf{x}$ is the sparse representation code and $\mathbf{D} \in \mathbb{R}^{m \times n}$ is called the equivalent dictionary matrix \cite{li2013projection} with $\mathbf{D} = \mathbf{A\Phi}$ and $m << n$. Another example is representation-based classification task where the dictionary is formed by collecting training samples column-wise and the aim is to represent a query sample by a linear combination of the columns (atoms) in the dictionary. In particular, this representation of the query sample $\mathbf{y}$ in the dictionary $\mathbf{D}$ is expressed as $\mathbf{y} = \mathbf{Dx}$, and solving it for $\mathbf{x}$ provides estimated representation coefficients $\widehat{\mathbf{x}}$. Then, the corresponding indices of the non-zero entries in $\widehat{\mathbf{x}}$ determine the predicted class label of the query sample. In this manner, there are different representation-based classification approaches consisting of Sparse Representation-based Classification (SRC) \cite{SRC1, SRC2} and Collaborative Representation-based Classification (CRC) \cite{collaborative}. In the first group, approaches are more focused on computing spare solutions $\widehat{\mathbf{x}}$, where query sample is represented sufficiently well with a few non-zero coefficients in the estimated solution. One major drawback of the SRC methods is that they are based on $\ell_1$-minimization requiring iterative computations which makes them computationally complex. In the CRC approach, the solution is computed by following the regularized least-square estimation as $\widehat{\mathbf{x}} = \left( \mathbf{D}^T \mathbf{D} + \lambda\mathbf{I} \right)^{-1}\mathbf{D}^T\mathbf{y}$ with the regularization parameter $\lambda$. A group selection procedure is used in the CRC method to determine the predicted label of the query sample. Estimated coefficients are replaced into the representation and the group with the least representation error is selected, therefore its corresponding class can be determined. Direct mapping for representation vector estimation is computationally more efficient than SRC approaches and in certain applications such as face recognition \cite{collaborative}, it can still provide comparable results.

Contrary to existing literature focusing widely on the SR task e.g., \cite{SRC1,SRC2,collaborative,reconnet,lamp,lista,lvamp}, there are only a few studies \cite{csen,csen-covid} concentrating on the SE problem. In fact, traditional methods for SE are based on first performing SR and then the support sets are estimated over the reconstructed signals by applying certain thresholding techniques. Considering the complexity of SR compared to the SE task, it is more feasible to directly estimate support sets. Moreover, in many applications, there might be no need to perform SR at all if the support sets are already obtained. For example, estimating the occupied spectrum in a CS task with cognitive radio systems \cite{SS1} is actually a SE task and it is substantially important as the spectrum interval is available only for a given time frame. Next, detection of an observed target in ground-penetrating radar systems \cite{radar1} is also a SE task since it only involves localization. Finally, in a representation-based classification task, it is important to locate active representation coefficients than their exact values since their locations solely determine the predicted class labels.

Because of the aforementioned reasons, for certain applications, it is more practical and efficient to develop the SE technique directly rather than a prior SR application. For this purpose, we proposed Convolutional Support Estimator Networks (CSENs) \cite{csen} combining traditional model-based SE with a learning-based approach. The CSENs are able to learn a mapping from a proxy signal which is a rough estimation for the sparse signal to be reconstructed, i.e., $\widetilde{\mathbf{x}} = \mathbf{By}$ where the denoiser matrix is defined as $\mathbf{B} = \mathbf{D}^ T$ or $\mathbf{B} = \left( \mathbf{D}^T\mathbf{D} + \lambda \mathbf{I} \right)^{-1}\mathbf{D}^T$. The CSENs can achieve state-of-the-art performance levels with minimum computational complexity. CSENs are designed to work with compact network architectures compared to deep SR networks such as ReconNet \cite{reconnet}, Learned Approximate Message Passing (LAMP) \cite{lamp}, Learned Iterative Shrinkage Thresholding Algorithm (LISTA) \cite{lista}, and Learned Vector AMP (LVAMP) \cite{lvamp}. Therefore, CSENs demonstrated that not only the compact network models are sufficient to learn a direct mapping for the support sets; they are also reliable over different measurement rates and robust against corruption in measurements, e.g., robust under different noise levels. Their extensions have been designed for COVID-19 classification \cite{csen-covid, csen-early-covid}, and recently for object distance estimation \cite{csen-distance} as a regressor.

However, especially in a compact configuration, the learning performance of CNNs is limited due to its homogenous network structure with a linear neuron model which also affected the performance of CSENs. Operational Neural Networks (ONNs) have recently been proposed \cite{kiranyaz2020operational, kiranyaz2021exploiting} as a superset of CNNs. ONNs have not only outperformed CNNs significantly, but they are even capable of learning those problems where CNNs entirely fail. On the other hand, ONNs like their ancestors, Generalized Operational Perceptrons (GOPs) \cite{kiranyaz2017progressive, tran2019heterogeneous} exhibited certain drawbacks such as strict dependability to the operators in the operator set for each layer/neuron, and the need for setting (fixing) the operator sets of the output layer neuron(s) in advance. Such drawbacks yield a limited network heterogeneity and divergence that eventually cause certain issues in learning performance and computational efficiency. As a solution, Self-organized ONNs (Self-ONNs) with the generative neuron model that can address all these drawbacks without any prior (operator) search or training, and with elegant computational complexity have been proposed \cite{onn1, onn2}. During the training of the network, to maximize the learning performance, each generative neuron in a Self-ONN can customize the nodal operators of each kernel connection. This yields an ultimate heterogeneity level that is far beyond what ONNs can offer, and thus, the traditional "weight optimization" of conventional CNNs is entirely turned out to be an "operator generation" process.

Traditional CNN architectures with compact configurations have another severe limitation which is the limited receptive field size \cite{receptive1, receptive2}. This is actually one of the main reasons for their limited learning capabilities. Similarly, compact Self-ONNs suffer from it due to the localized and fixed-size kernels. To address this drawback, a recent study in \cite{super1} has introduced superior generative neurons, or "Super Neurons" in short that can increase the receptive field size substantially with non-localized kernels. Considering the fact that kernel locations are jointly optimized along with the kernel parameters, each super neuron is able to learn the best kernel transformation function for the kernel position optimized simultaneously. Hence, during the inference stage, super neurons have approximately similar computational complexity with only additional shifting operations and kernel shift parameters. It is shown in \cite{super1} that compact Self-ONNs with super neurons can achieve significantly higher performance levels in several regression and classification problems.

The CSENs have naturally inherited the aforementioned limitations of compact CNNs, limited learning ability using linear transformation functions and fixed kernel locations. Furthermore, their training objective in representation-based classification problems lacks the contribution from corresponding SE tasks as will be discussed in the following section. Finally, linear proxy mapping operation plays an additional limiting factor for the initial estimation. To address these drawbacks, in this study, we propose novel \textit{Operational Support Estimator Networks (OSENs)} based on the operational layers of Self-ONNs with super neurons. OSENs can learn a direct mapping for the SE and can significantly outperform the state-of-the-art CSENs and other traditional SR-based approaches. The novel and significant contributions of this work can be summarized as follows:
\begin{itemize}
    \item We propose a novel OSEN approach for a non-iterative SE. The proposed approach has achieved state-of-the-art performance levels in simulated SE problems over the MNIST dataset. Moreover, a hybrid loss penalizing both SE estimation and classification errors is proposed in the representation-based classification framework using OSENs. It has achieved superior classification accuracy on Yale-B dataset.

    \item A Non-linear Compressive Learning (NCL) approach is proposed for joint optimization of the proxy mapping and SE parts. Consequently, NCL-OSENs can directly compute a mapping from measurement signals to support locations, whereas conventional CSENs need a proxy computation for their input: $\widetilde{\mathbf{x}} = \mathbf{By}$.

    \item We introduce Self-Organized Generalized Operational Perceptrons (Self-GOPs) in the NCL module of the proposed approach. This is the first study in the literature introducing self-organizing GOP models in fully connected layers. The Self-GOPs can learn non-linear mappings in dense layers and it is shown that their learning capability is superior compared to conventional Multi-Layer Perceptrons (MLPs).

    \item OSENs produce probability maps indicating the probability of signal coefficients being non-zero in a sparse representation vector. This prior likelihood can be used as prior information in signal reconstruction algorithms to further enhance CS performance. We show that the reconstruction accuracy is improved using proposed \textit{learning-aided} recovery framework in Magnetic Resonance Imaging (MRI) CS problem with a total-variation (TV) based minimization scheme.

    \item Thanks to the operational layers with super neurons, compact OSENs have an elegant learning capability with a limited amount of training data. Furthermore, it is computationally efficient because of more compact and shallow network architectures compared to the competing methods.
    
    \item Finally, the number of studies focusing on solely SE is only a few compared to the well-studied SR applications. Hence, this work with state-of-the-art performance levels achieved in numerous applications will stir the attention and eventually draw more interest in SE.
\end{itemize}

The rest of the paper is organized as follows, background and prior work are provided in Section \ref{background}, then the proposed methodology will be presented in Section \ref{method}, and experimental setup and an extensive set of comparative evaluations for different applications have been presented in Section \ref{result}. Finally, Section \ref{conclusion} concludes the paper and suggests topics for future research.

\section{Background and Prior Work}
\label{background}

In this section, we first introduce the notations that are used in this study and provide a brief background related to sparse representation, SE, and representation-based classification.

The $\ell_p$-norm of a vector $\mathbf{x} \in \mathbb{R}^n$ is defined as $\left \| \mathbf{x} \right \|_{\ell_p^n} =   \left (  \sum_{i=1}^n \left \vert x_i \right \vert^p \right )^{1/p}$ for $p \geq 1$, whereas the $\ell_0$-norm is $\left \| \mathbf{x} \right \|_{\ell_0^n} = \lim_{p \to 0} \sum_{i=1}^n \left \vert x_i \right \vert^p = \# \{ j: x_j \neq 0 \}$ and $\ell_{\infty}$-norm is $\left \| \mathbf{x} \right \|_{\ell_{\infty}^n} =  \max_{i=1,...,n} \left (   \left | x_i \right | \right )$. A signal $\mathbf{s}$ is called strictly $k$-sparse if it can be represented in a proper domain, i.e., $\mathbf{s}= \mathbf{ \Phi}\ \mathbf{x}$ by less than $k+1$ non-zero representation coefficients such that $\left \|  \mathbf{x} \right \|_0 \leq k$. The location information of these non-zero coefficients forms the support set $\Lambda := \left \{ i:  x_i \neq 0 \right \}$. In other words, $\Lambda \subset \{1,2,3,...,n \}$ is a set of indices corresponding to active basis vectors in the representation of the signal $\mathbf{s}$ in the domain $\mathbf{ \Phi}$.

In CS theory, a signal $\mathbf{s}$ is sensed with a few number of measurements,
 \begin{equation}
     \mathbf{y} = \mathbf{A} \mathbf{s} = \mathbf{A}\mathbf{ \Phi} \mathbf{x} =\mathbf{ D} \mathbf{x}
     \label{Eq:CS},
 \end{equation}
where $\mathbf{A} \in \mathbb{R}^{m \times d}$ and $\mathbf{D} \in \mathbb{R}^{m \times n}$ are called measurement matrix and equivalent dictionary, respectively, and we define the measurement rate (MR) as $m/n$. This system is an underdetermined linear system of equations as $m < < n$; therefore, a priori assumption about the solution is needed to find a unique solution for this ill-posed problem. It is shown in \cite{spark} that the sparse representation of $\mathbf{x}$ satisfying $\left \| \mathbf{x }\right \|_{0} \leq k$ is unique in the solution,
\begin{equation}
\min_\mathbf{x} ~ \left \| \mathbf{x }\right \|_{0}~ \text{subject to}~ \mathbf{D} \mathbf{x} = \mathbf{y}, \label{Eq:l0}
\end{equation}
if $\mathbf{D}$ has more than $2k$ linearly independent columns and $m \geq 2k$. That is to say, at least $k$-sparse signal pairs are distinguishable or they can be separately represented in the dictionary $\mathbf{D}$. However, the problem in \eqref{Eq:l0} is NP-hard and non-convex because of $\ell_0$-norm. Its closest norm relaxation can be the following so-called Basis Pursuit \cite{BP}:
\begin{equation}
     \min_\mathbf{x} \left \| \mathbf{x} \right \|_1 ~\text{s.t.} ~ \mathbf{x} \in \mho \left (\mathbf{ y} \right ), \label{Eq:l1}
\end{equation}
where $\mho \left ( \mathbf{y} \right ) = \left \{ \mathbf{x}: \mathbf{D} \mathbf{x}=\mathbf{y} \right \}$. The CS theory claims that in cases where the exact recovery of $\mathbf{s}$ is not possible, a tractable solution is still achievable if $m>k(log(n/k))$ and Restricted Isometry Property \cite{candesRIP} is satisfied for $\mathbf{D}$. If these conditions are satisfied, the stable recovery from corrupted noisy query sample is also possible via Basis Pursuit Denoising (BPDN) \cite{BP} using a relaxed version of \eqref{Eq:l1}, i.e., $\min_{\mathbf{x}} \left \| \mathbf{x} \right \| ~\text{s.t.} ~    \left \| \mathbf{y} - \mathbf{Dx} \right \|  \leq \epsilon$, where a small $\epsilon$ constant is set according to the noise power.

\subsection{Sparse Support Estimation (SE)}

In various applications, the SE task is more important than performing a complete SR. The complete SR procedure includes finding the support set, magnitude, and their corresponding signs. For example, in an anomaly detection problem using distributed CS surveillance systems, location of non-zero indices (support set $\lambda$) is satisfactory enough to locate the anomaly such as anomaly detection in sensor data streams \cite{anomaly1} and HSI images \cite{anomaly2}. Similarly, active user detection in NOMA \cite{Noma1, Noma2} and CDMA \cite{CDMA} systems can be considered SE tasks with a significant role in 5G communication. Additionally, considering a query sample in a representation-based classification problem \cite{SRC1,SRC2,collaborative}, finding the support set already provides class information of the query sample; hence, there is no need to perform computationally expensive estimation for the exact $\mathbf{x}$.

A support estimator $\mathcal{E}(.)$ is defined as follows,
\begin{equation}
   \widehat{\Lambda} =  \mathcal{E}\left (\mathbf{y},\mathbf{D} \right ), 
\end{equation}
where $\mathbf{y}= \mathbf{Dx} +\mathbf{z}$ is the measurement with an additive noise and $\widehat{\Lambda}$ is the estimated support set. Previously, traditional SE methods were based on recovering the exact signal which makes their estimation performance dependent on the SR. These conventional methods can be grouped under three categories: i. iterative estimators using $\ell_1$-minimization \cite{SRC1,SRC2,fast,l1ls,ADMM,homotopy,gpsr,l1magic}, ii. least-square sense approximation including LMSEE \cite{AMP-partial} and Maximum Correlation (MC) \cite{MaximumCorrelation} as $\widehat{\mathbf{x}}^{\text{LMMSE}} = \left ( \mathbf{D}^T \mathbf{D} + \lambda \mathbf{I}_{n \times n} \right )^{-1} \mathbf{D}^T \mathbf{ y}$ and $\widehat{\mathbf{x}}^{\text{MC}} = \mathbf{D}^T \mathbf{y}$, respectively, and iii. Deep Neural Networks \cite{reconnet,lamp,lista,lvamp}.

The approaches in (i) are computationally complex and iterative methods limiting their efficiency in support recovery. Although the approaches in (ii) use the closed-form solution and are non-iterative, their accuracies might be limited in challenging cases 
 such as low MRs \cite{AMP-partial}. In the last group (iii), various deep learning approaches are proposed mainly for SR. Deep unfolding models in \cite{lamp,lista,lvamp} consist of many dense layers with high numbers of trainable parameters in millions; ReconNet in \cite{reconnet} has several fully convolutional layers with a deep network configuration. Overall, these methods aim to compute a direct mapping from a measurement to the original signal using complex architectures with many layers. Therefore, their usability is limited when available training data is scarce. Their generalization capability is only satisfied with massive sizes of data and it is shown in \cite{csen} that they are also sensitive to noisy measurements.

 \subsection{Convolutional Support Estimator Networks (CSENs)}

In our recent work, CSENs \cite{csen} are proposed for the SE task as a non-iterative and computationally efficient approach. These networks are designed to compute support sets without performing a prior SR. It is shown that CSENs can outperform the traditional methods which often produce noisy estimations due to the uncertainty in the prior recovery. Thanks to their proposed compact network configuration, it is possible to achieve satisfactory performance levels using limited training data. Furthermore, the compact architecture provides improved generalization capability and further enhances the robustness of CSENs to measurement noises. The readers are referred to \cite{csen} for more detailed evaluations and technical discussions where limitations of classical support estimators are discussed and compared against non-iterative network-based estimators.

The proposed approach in this study has the following novelties and improvements over the previous state-of-the-art CSENs. First, CSENs are based on compact configurations using linear neuron models whose performance is limited compared to the proposed OSENs that can learn to approximate non-linear kernel transformation functions. Secondly, the previous network had fixed-size receptive fields, whereas in this study, we propose to follow non-localized kernels by jointly optimizing the kernel locations with the support estimator. Next, we introduce a hybrid loss penalizing both representation-based classification and support estimation errors, while the CSENs were trained using solely classification loss for the representation-based classification problem. Moreover, we propose the NCL module integrated into the OSENs for non-linear compressive learning to improve the support estimation and classification performances. Finally, the OSEN approach is evaluated in a CS-MRI task with the proposed learning-aided CS framework where we formulate the NCL module for the complex-domain non-linear proxy mapping.

\subsection{Representation-based Classification}

As introduced earlier, traditional approaches proposed for representation-based classification can be grouped into two categories: SRC and CRC methods. In both categories, the estimation of the support set is the main goal. Generally, a dictionary is built by stacking training samples column-wise and when a test sample is introduced, it is aimed to represent the test sample by a linear combination of the columns of representative dictionary $\mathbf{D}$. Predicted label for the test sample is assigned according to the location of the estimated non-zero representation coefficients. Specifically, it is expected that the solution vector $\widehat{\mathbf{x}}$ has sufficient information to represent query sample $\mathbf{y} = \mathbf{D} \widehat{\mathbf{x}} + \mathbf{z}$ within a small error margin and the query sample has the same label as the dictionary samples whose corresponding coefficients are non-zero.

\subsubsection{Sparse Representation-based Classification (SRC)}
There are various existing SRC approaches that are proposed for different applications, for example, early coronavirus disease 2019 (COVID-19) detection \cite{csen-early-covid}, COVID-19 recognition \cite{csen-covid}, hyper-spectral image classification \cite{hyperspecral}, face recognition \cite{SRC2}, and human action recognition \cite{human-action}. Generally, these approaches try to represent query sample $\mathbf{y}$ using only a few coefficients. It is expected that these non-zero components of $\widehat{\mathbf{x}}$ should have the same class label as the query. More specifically, the recovery of sparse signal is obtained using, e.g., Lasso formulation \cite{lasso-stable} for $\mathbf{y} =\mathbf{D} \mathbf{x} + \mathbf{z}$:
\begin{equation}
\label{eq:lasso}
    \min_\mathbf{x}  \left \{ \left \|  \mathbf{D}\mathbf{x}-\mathbf{y} \right \|_2^2 + \lambda \left \|\mathbf{x} \right \|_1  \right \},   
\end{equation}
where $\kappa$ is a small constant and for a stable solution, it satisfies that $\left \| \mathbf{x}- \widehat{\mathbf{x}} \right \| \leq \kappa \left \| \mathbf{z} \right \|$. This $\ell_1$-minimization is used in a four-step classification procedure consisting of (i) normalization of $\mathbf{y}$ and $\mathbf{D}$ to have unit $\ell_2$-norm, (ii) estimation of $\widehat{\mathbf{x}} = \arg \min_{\mathbf{x}} \left \|\mathbf{ x} \right \|_1 \text{s.t} \left \| \mathbf{y} - \mathbf{D} \mathbf{x} \right \|_2$, (iii) computing $\mathbf{e}_i = \left \| \mathbf{y} - \mathbf{D}_i  \widehat{\mathbf{x}}_i \right \|_2$, where $\mathbf{e}_i$ is residual for class $i$ and $\widehat{\mathbf{x}}_i$ is the estimated group coefficients, and (iv) label is assigned to the query sample by $\text{Class}\left ( \mathbf{y} \right ) = \arg \min \left ( \mathbf{e}_i \right )$. The four-step approach or its similar variants is widely used in the abovementioned SRC studies \cite{csen-covid, csen-early-covid, human-action, hyperspecral}. This residual finding approach is more suitable for real-life classification problems where there is an existing correlation between samples. 

\subsubsection{Collaborative Representation-based Classification (CRC)}

Although SRC approaches achieve considerable performance levels, they have one major limitation that $\ell_1$-minimization in their SR scheme requires iterative computation increasing their computational complexity. To address this drawback, the CRC approach in \cite{collaborative} proposes that instead of using \eqref{eq:lasso}, traditional $\ell_2$-based minimization can be followed,
\begin{equation}
    \label{eq:l2}
    \mathbf{\widehat{x} }= \arg \min_{\mathbf{x}} \left \{ \left \| \mathbf{y }- \mathbf{D}\mathbf{x }\right \|_2^2  + \lambda \left \| \mathbf{x} \right \|_2^2   \right \}.
\end{equation}
The above problem has a closed-form solution, i.e., $\widehat{\mathbf{x}} = \left ( \mathbf{D}^T  \mathbf{D} + \lambda \mathbf{I}_{n \times n}  \right )^{-1} \mathbf{D}^T  \mathbf{ y}$, which can be replaced in the second step of the four-step classification approach.

The CRC approach focuses on finding the minimum energy solution instead of computing the sparsest $\widehat{\mathbf{x}}$. Hence, query signal $\mathbf{y}$ is represented using relatively small coefficients where a collaborative representation is searched within the atoms of the dictionary. It is discussed in \cite{collaborative} that such collaborative representation might be preferred in the cases where dictionary $\mathbf{D}$ is unable to satisfy the properties required by the exact or stable recovery due to correlation between images/signals. Especially when MR is large enough, $\ell_2$-minimization provides similar or even better recognition performances compared to SRC approaches. Additionally, the CRC approach is computationally efficient since it computes a direct-mapping by the closed-form solution.

\begin{figure*}[ht]
    \centering
    \includegraphics[width=\linewidth]{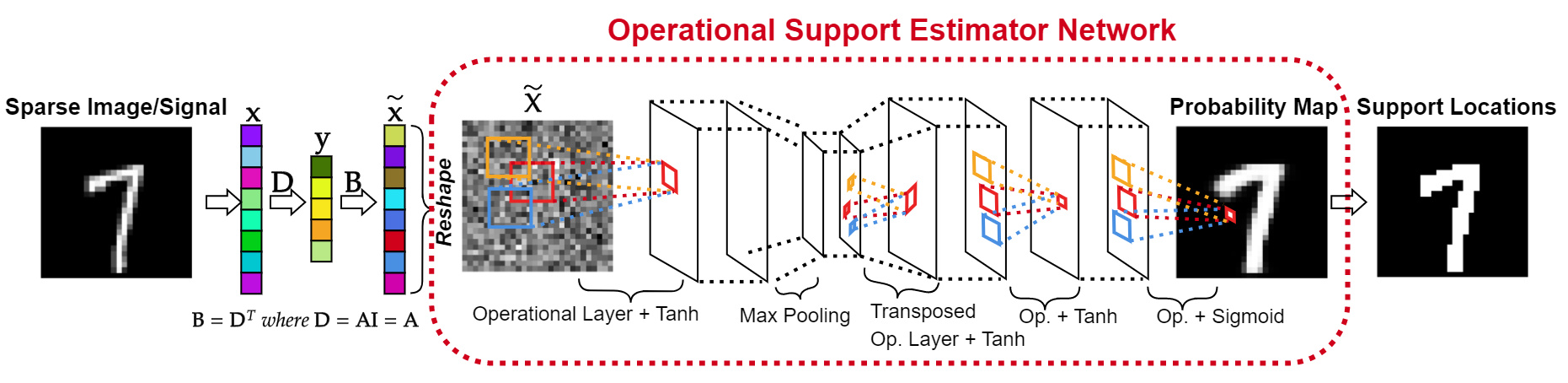}
    \caption{The proposed OSEN approach with non-localized and non-linear kernels for the SE task. For a sparse image/signal $\mathbf{x}$ in spatial domain ($\mathbf{\Phi} = \mathbf{I}$), rough and noisy estimation is first performed by $\widetilde{\mathbf{x}}=\mathbf{B}\mathbf{y}$ where $\mathbf{y}$ is the measurement. Support locations corresponding to non-zero elements are then found in the form of segmentation masks.}
\label{fig:osen}
\end{figure*}

\section{Proposed Methodology}
\label{method}

In this section, we will introduce proposed OSENs equipped with self-organized operational layers and NCL-OSEN approach where we jointly optimize the compression matrix with support estimator in the training stage. Finally, a learning-aided SR technique is presented where the output of the OSENs is used as side information in exact SR tasks.

\subsection{Operational Support Estimator Networks (OSENs)}

Given $\mathbf{y}$ and $\mathbf{D}$, the proposed SE approach learns to produce a binary mask $\mathbf{v} \in \left \{ 0,1 \right \}^n$ indicating active support elements:
\begin{subnumcases}
{v_i =}
   1    \hfill & \text{ if $ i \in \Lambda, $ } \label{binary1} \\
   0 & \text{ else }. \label{binary2}
\end{subnumcases}
The estimation of such binary mask, $\widehat{\mathbf{v}}$, is equivalent to finding the support set as $\widehat{\Lambda} = \left \{  i \in  \left \{  1,2,..,n\right \} : \widehat{v}_i =1   \right \}$. In this manner, we train the OSEN network for this segmentation problem to provide the following mapping: $\mathcal{P} \left ( \widetilde{\mathbf{x}}  \right ): \mathbb{R}^n  \mapsto \left [ 0,1 \right ]^n$, where $\widetilde{\mathbf{x}}$ is called proxy estimation which is computed by LMMSE as $\widetilde{\mathbf{x}} = \left ( \mathbf{D}^T \mathbf{D} + \lambda \mathbf{I} \right )^{-1} \mathbf{D}^T\mathbf{y}$ or MC $\widetilde{\mathbf{x}} = \mathbf{D}^T\mathbf{y}$. Then, the proxy is reshaped to a 2D plane and fed as the input to the OSEN.

The proposed OSEN approach consists of self-organized operational layers where non-linear kernel transformation function of each super neuron is approximated using Taylor series expansion. Accordingly, the $Q^{\text{th}}$ order approximation near the origin can be written as a finite sum of derivatives for a function $g$ at given point $x$ as follows,
\begin{equation}
    g(x)^{(Q)} = \sum_{q=0}^{Q}\frac{g^{(q)}(0)}{q!}(x)^q.
\end{equation}
Assuming $w_q = \frac{g^{(q)}(0)}{q!}$ as the $q^{th}$ coefficient of the approximation, the set of coefficients $\mathbf{w} \in \mathbb{R}^Q$ is learned during training. For a given single channel input $\mathbf{x}^{(k)} \in \mathbb{R} ^ {P \times R}$, the $k^{\text{th}}$ filter in an operational layer having shared weights would have the following set of trainable parameters: $\mathbf{W}^{(k)} = [\mathbf{w}_1^{(k)}, \mathbf{w}_2^{(k)}, \dots, \mathbf{w}_Q^{(k)}] \in \mathbb{R}^{f_s \times f_s \times Q}$, where $\mathbf{w}_q^{(k)} \in \mathbb{R}^{f_s \times f_s}$ is the $q^{\text{th}}$ coefficient and $f_s$ is the filter size.

\begin{figure*}[ht]
    \centering
    \includegraphics[width=0.9\linewidth]{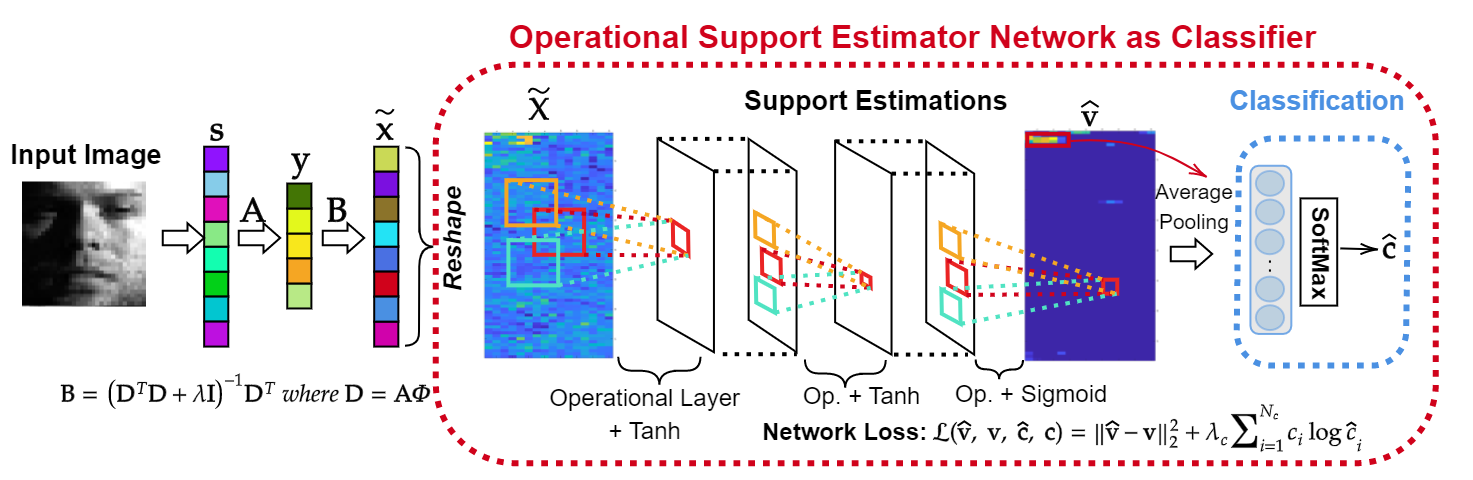}
    \caption{The proposed OSEN approach with non-localized and non-linear kernels for the representation-based classification task. The representative dictionary $\mathbf{D}$ is formed by applying a dimensionality reduction with matrix $\mathbf{A}$ to stacked dictionary samples ($\mathbf{\Phi}$). For a vectorized query image $\mathbf{y}$, rough and noisy estimation for support locations is performed by $\widetilde{\mathbf{x}}=\mathbf{B}\mathbf{y}$. Then, support locations are computed to estimate the class label.}     
\label{fig:osen_class}
\end{figure*}

The intermediate output of the $k^\text{th}$ filter at pixel $(p, r)$ would be
\begin{equation}
\label{eq:intermediate}
    {g(\mathbf{x}^{(k)})}_{p, r} = \sum_{q=0}^Q \sum_{i=0}^{f_s-1} \sum_{j=0}^{f_s-1} w_{q, i, j}^{(k)}\left(x_{p + i + \alpha, r + j + \beta}^{(k)} \right)^q,
\end{equation}
where $\alpha$ and $\beta$ are the shift (bias) parameters optimized during training. The summations are commutative in \eqref{eq:intermediate}, then more computationally efficient implementation can be desired by expressing the operational layers as the summation of repeated convolutions,
\begin{equation}
    \label{eq:operation}
    \widehat{\mathbf{x}}^{(k)} = \sigma \left( \sum_{q=1}^Q \left(\mathbf{x}^{(k)}_t \right)^{\odot q} * \mathbf{w}_q^{(k)} + b_q^{(k)} \right),
\end{equation}
where $*$ is the convolution operation, $\mathbf{x}^{\odot q}$ is the Hadamard power providing component-wise power raise $x \mapsto x^q$, $b_q^{(k)}$ is bias, and $\widehat{\mathbf{x}}^{(k)}$ is the final output of the generative neuron after activation function $\sigma(.)$ is applied. Note that $\mathbf{x}^{(k)}_t$ is obtained by shifting the input such that $\mathbf{x}^{(k)}_t = T^{(k)}_{\alpha, \beta}\mathbf{x}^{(k)}$ where $T^{(k)}_{\alpha, \beta}$ is the shifting operator for the $k^\text{th}$ kernel. In this implementation, we use real-valued shifts $\alpha, \beta \in \mathbb{R}$, where the shifted feature maps are computed by bilinear interpolation.

Overall, the $k^\text{th}$ feature map of $l^\text{th}$ layer is computed using the following trainable parameters $ \mathbf{\Theta}^{(k)}_l = \big \{\mathbf{W}^{(k)}_l \in \mathbb{R}^{f_s \times f_s \times Q \times N_{l-1}}, \mathbf{b}^{(k)}_l \in \mathbb{R}^{Q}, \alpha^{(k)}_l \in \mathbb{R}, \beta^{(k)}_l \in \mathbb{R} \big \}$ where $N_{l-1}$ is the number of feature maps in the previous layer. As $\alpha_l,\beta_l$ are scalar values, shift parameters are shared for the previous layer connections of the $k^\text{th}$ feature map. Finally, total trainable parameters in a $L$-layer OSEN will be $\mathbf{\Theta}_\text{OSEN}=\big\{ \{\mathbf{\Theta}^{(k)}_1\}_{k=1}^{N_1}, \{\mathbf{\Theta}^{(k)}_2\}_{k=1}^{N_2}, ... ,  \{\mathbf{\Theta}^{(k)}_L\}_{k=1}^{N_L}\big\}$. Shifting parameters of the kernels are jointly learned with kernel parameters, which determine the transformation function of each kernel element. 

During training, Mean-Square Error (MSE) is computed for training sample pairs of $(\widetilde{\mathbf{x}}, \mathbf{v})$ as,
\begin{equation}
\label{eq:mse_cost}
\mathcal{L}(\mathbf{\Theta}_\text{OSEN}, \widetilde{\mathbf{x}}, \mathbf{v}) = \left \| \mathcal{P}_{\mathbf{\Theta}_\text{OSEN}}\left (\widetilde{\mathbf{x}} \right )- \mathbf{v} \right \|_2^{2}.
\end{equation}
In specific to representation-based classification, instead of using \eqref{eq:mse_cost}, a group $\ell_2$-minimization based optimization can be followed:
\begin{equation}
\label{eq:rbr_cost}
\mathcal{L_G}(\widehat{\mathbf{v}}, \mathbf{v}) = \left \| \widehat{\mathbf{v}} - \mathbf{v} \right \|_2^{2} + \lambda_g \sum_{i=1}^{N_c}\left \| \widehat{\mathbf{v}}_{G,i} \right \|_2. 
\end{equation}
where $\widehat{\mathbf{v}} = \mathcal{P}_{\mathbf{\Theta}_\text{OSEN}}\left (\widetilde{\mathbf{x}} \right )$ is the predicted mask and $\widehat{\mathbf{v}}_{G,i}$ is the group of supports for class $i$. In the previous work \cite{csen}, this cost function is approximated by average-pooling which is followed by a SoftMax operation at the output. Hence, it was possible to produce class probabilities directly from the input proxy. On the other hand, such an approximation may reduce the classification performance since the MSE between the predicted and actual binary masks indicating support sets is ignored for the training sample pairs. Therefore, in this study, we propose a novel approach that can produce support maps and estimated class labels after a single inference. Accordingly, the proposed approach provides the following mapping: $\mathcal{P}_{\mathbf{\Theta}_\text{OSEN}}\left (\widetilde{\mathbf{x}} \right ) = \left\{ \widehat{\mathbf{v}}, \widehat{\mathbf{c}}_y \right\}$, where $\widehat{\mathbf{c}}_y \in \mathbb{R}^{N_C}$ is the estimated class label vector for the query sample $\mathbf{y}$. The following hybrid loss function is proposed to train the network:
\begin{equation}
\label{eq:hyb_rbr_cost}
\mathcal{L_R}(\widehat{\mathbf{v}}, \mathbf{v}, \widehat{\mathbf{c}}_y, \mathbf{c}_y) = \left \| \widehat{\mathbf{v}} - \mathbf{v} \right \|_2^{2} + \lambda_c \sum_{i=1}^{N_C} c_{y,i} \log(\widehat{c}_{y,i}). 
\end{equation}
In this way, the estimated support masks and class labels are optimized jointly during the training of OSENs in representation-based classification task.

Overall, if SE problem is different than the classification, the OSEN model is trained using \eqref{eq:mse_cost} as its cost function with the following input-output pairs: $(\widetilde{\mathbf{x}}^\text{train}, \mathbf{v}^\text{train})$ as illustrated in Fig. \ref{fig:osen}. If there is categorical class label information available, then it is trained using $(\widetilde{\mathbf{x}}^\text{train}, \mathbf{v}^\text{train}, \mathbf{c}^\text{train}_y)$ triplet samples with the hybrid loss in \eqref{eq:hyb_rbr_cost} as presented in Fig. \ref{fig:osen_class}.

\subsection{Non-linear Compressive Learning (NCL)}

The OSEN approach takes proxy $\widetilde{\mathbf{x}} = \mathbf{B}\mathbf{y}$ signal as the input, where $\mathbf{B}$ is the denoiser matrix according to LMMSE or MC. This rough estimation may limit the potential of non-linear neuron model in the proposed approach. Let us consider a measurement $\mathbf{y}$, and we will attempt to recover $\mathbf{x}$ from $\mathbf{y} = \mathbf{D}\mathbf{x}$. Using the Fundamental Theorem of Linear Algebra \cite{strang1993fundamental}, for $m < n$, we can see that the solution set contains an infinite number of elements, as $\mathbf{D}$ has a null space with a dimensionality of at least $n-m$. Therefore, in order to satisfy the uniqueness of the solution, we first need the eliminate the null space and map $\mathbf{y}$ to the row space of $\mathbf{D}$. For simplicity, let us also assume $\mathbf{D}$ is full row rank. Even for this case, there is no linear mapping that can guarantee the exact recovery of $\mathbf{x}$. Moreover, if we want to use linear mapping, such linear mapping from the column space to the row space is not unique. For instance, both $\mathbf{B} = \mathbf{D}^T$ and $\mathbf{B}=\left( \mathbf{D}^T\mathbf{D}+\lambda\mathbf{I} \right)^{-1}\mathbf{D}^T$ are linear mapping from column space to row space and each produces a unique solution for full row rank $\mathbf{D}$. Different values of $\lambda$ result in infinitely many such linear mappings to select from. Therefore, using a fully connected layer (i.e., an MLP) can learn such linear mapping instead of relying on hand-crafted linear mappings mentioned above.

On the other hand, uniqueness does not guarantee, exact recovery. The exact recovery can only be guaranteed for a special subspace of row space, which is set of all $k$-sparse signals. However, in this case, the mapping cannot be linear; it must be non-linear. Theoretically, one might attempt to try all possible support sets by selecting $k$ atoms from $\mathbf{D}$ and then applying linear mapping to each, but this approach is NP-hard. Indeed, in practice, such mappings are realized by iteratively employing linear mappings, each followed by shrinkage operations. To address this drawback, we propose the NCL module that can learn as close as possible approximation of such ideal iterative mapping, with a single-layer, non-iterative, and learned non-linear operator. In the proposed model, we jointly optimize proxy-mapping stage with the support estimator part.

\begin{figure}[h]
    \centering
    \includegraphics[width=\linewidth]{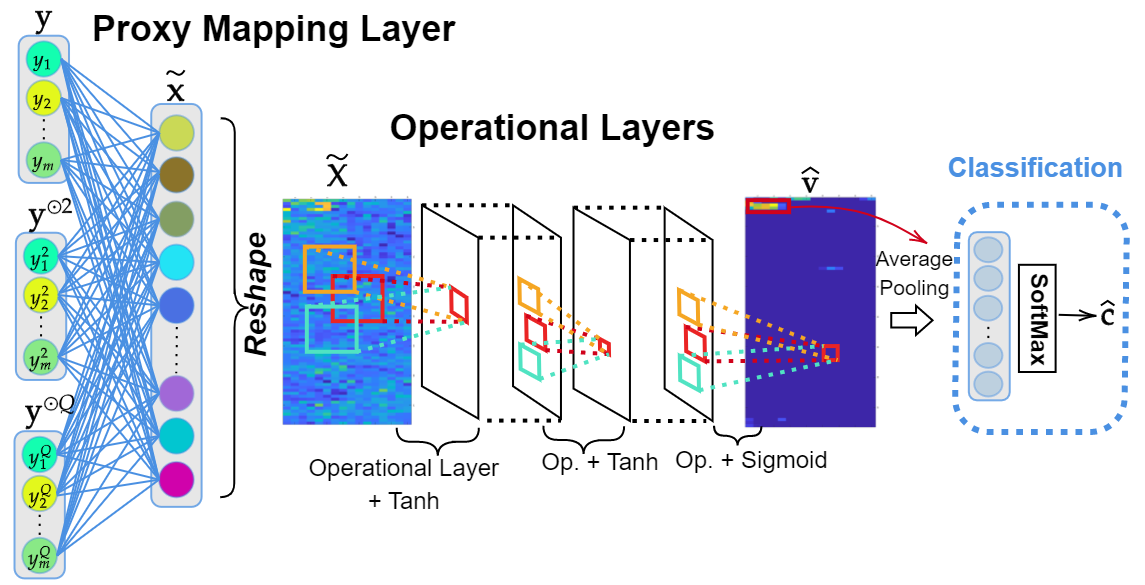}
    \caption{The proposed NCL-OSEN approach has a proxy mapping layer equipped with Self-GOPs as defined in \eqref{eq:perceptron}. The network is trained end-to-end and the proxy mapping layer is jointly optimized with the support estimation and classification parts.}     
\label{fig:ncl_osen}
\end{figure}

We introduce the Self-GOP model with the \textit{generative} neuron topology to approximate different transformation functions that are not possible to learn with a traditional MLP approach. Similar to operational layers, a Self-GOP layer can learn highly non-linear mappings with less number of neurons. These neurons are named as \textit{generative} perceptrons in Self-GOP model since they are formed to perform the following transformation during the training process:
\begin{equation}
    g(x, \mathbf{w}) = w_0 + xw_1 + x^2w_2 + \dots + x^Qw_Q,
\end{equation}
where $\mathbf{w} \in \mathbb{R}^Q$ is the trainable weight for a scalar input $x$. The proposed NCL scheme has one fully-connected layer consisting of Self-GOPs. Given a measurement signal $\mathbf{y}$ as input, it tries to learn the following mapping, $\phi(\mathbf{y}): \mathbb{R}^m \mapsto \mathbb{R}^n$ for $n > m$. Since summations are commutative, one can write the output for input $\mathbf{y}$ as,
\begin{equation}
    \label{eq:perceptron}
    \widetilde{\mathbf{x}} = \phi(\mathbf{y}) = \sigma \left( \sum_{q=1}^Q \mathbf{w}_q \mathbf{y}^{\odot q} + \mathbf{b}_q \right),
\end{equation}
where $\mathbf{w}_q \in \mathbb{R}^{n \times m}$ and $\mathbf{b}_q \in \mathbb{R}^n$. Then, overall trainable parameters would be: $ \mathbf{\Theta}_\text{NCL} = \big \{\mathbf{W} \in \mathbb{R}^{n \times m \times Q}, \mathbf{b} \in \mathbb{R}^{n \times Q} \big \}$. Note that the first order weights in \eqref{eq:perceptron} are initialized with $\mathbf{w}_1 = \mathbf{B}$ where $\mathbf{B}=\left( \mathbf{D}^T\mathbf{D}+\lambda\mathbf{I} \right)^{-1}\mathbf{D}^T$ for the representation-based classification problem and $\mathbf{B} = \mathbf{D}^T$ for other SE task. Finally, the output of the proposed NCL scheme is connected to the OSEN. Therefore, output $\widetilde{\mathbf{x}} = \phi(\mathbf{y})$ is reshaped to be fed to the first operational layer of the OSEN.

In this work, we name this end-to-end novel approach as NCL-OSENs where the proxy mapping from low to high-dimensional space is learned using Self-GOPs during training. It is observed that the joint optimization of $\mathbf{\Theta}_\text{NCL}$ and $\mathbf{\Theta}_\text{OSEN}$ improves SE performance. As illustrated in Fig. \ref{fig:ncl_osen}, the input-output training pairs of the NCL-OSEN approach would be $( \mathbf{y}^\text{train}, \mathbf{v}^\text{train})$ and in case of classification it will be $( \mathbf{y}^\text{train}, \mathbf{v}^\text{train}, c_y^\text{train})$. Note the fact that the NCL module of NCL-OSEN for $Q=1$ configuration is actually equivalent to having MLP as a proxy mapping layer with a tangent hyperbolic activation function.

\subsection{Learning-aided Signal Reconstruction via Total-Variation Minimization}

When the aim is to recover the exact signal, produced probability maps by OSENs can be used as side information in $\ell_1$-minimization based approaches. In this case, the following weighted minimization approach can be followed:
\begin{equation}
\label{eq:weighted-lasso}
    \min_\mathbf{x}  \left \{ \left \|  \mathbf{D}\mathbf{x}-\mathbf{y} \right \|_2^2 + \lambda \left \| \mathbf{\Gamma} \odot \mathbf{x} \right \|_1  \right \},
\end{equation}
where $\mathbf{\Gamma} \in \mathbb{R}^n$ is the weight for non-zero cost and $\odot$ is the component-wise multiplication operator. The weight is formed by prior information about each component of the sparse signal $\mathbf{x}$. In modified-CS literature \cite{Modified-CS}, the weight of the $i^\text{th}$ element is defined as $\gamma_i = \frac{1}{p_{i} + \epsilon}$, where $\epsilon$ is a small positive constant and $p_{i}$ is the $i^\text{th}$ element of $\mathbf{p}$ that is e.g., prior likelihood \cite{priori1} of $\Lambda$ such that the probability of $x_i$ being non-zero is $p_i$.

In the learning-aided CS framework, we choose gradient as the sparsifying domain: $\mathbf{\Phi} = \mathbf{\nabla}$ in the aforementioned CS scheme, i.e., $\mathbf{y} = \mathbf{A} \mathbf{s} = \mathbf{A}\mathbf{ \Phi} \mathbf{x} =\mathbf{ D} \mathbf{x}$. The gradient-domain is a convenient choice since natural images/signals are generally sparse in $\mathbf{\nabla}$, which can also preserve boundary details and edges as a preferred way of reconstruction in many inverse imaging systems \cite{comp_book, tv1, tv2}. Let $\mathbf{S} \in \mathbb{R}^{P \times R}$ be an image that will be compressively sensed, then the following total-variation minimization problem is defined:
\begin{equation}
    \min_{\mathbf{S}} \left \{ \left \| \mathbf{y} -\mathbf{A} \text{vec}(\mathbf{S}) \right \|_2^2 + \lambda  \left \| \mathbf{\nabla} \mathbf{S} \right \|_\text{TV} \right \},
    \label{eq:tv}
\end{equation}
where $\text{vec}(.)$ is the vectorization operation and $\left \| \nabla \mathbf{S} \right \|_\text{TV}$ is defined as follows,
\begin{align}
     \left \| \nabla \mathbf{S} \right \|_\text{TV} & =     \left \| \nabla_x \mathbf{S} \right \|_1 + \left \| \nabla_y \mathbf{S} \right \|_1 \\ &= \sum_{p,r} \left | S_{p+1,r} - S_{p,r}\right | + \sum_{p,r} \left | S_{p,r+1} - S_{p,r}\right |,
\end{align}
i.e., the anisotropic total-variation. We solve the problem in \eqref{eq:tv} using Total-Variation (TV) minimizer by Alternating Direction Method of Multipliers (ADMM) \cite{TVAL3, ADMM}. 

The OSEN approach for learning-aided SR has two input and output channels. Given a measurement $\mathbf{y} = \mathbf{A}\text{vec}(\mathbf{S})$, first rough estimation $\widetilde{\mathbf{S}} = \mathbf{A}^T\mathbf{y}$ is computed, then it is reshaped to the 2-D plane. Consequently, the training input samples consisting of two-channel proxy will be $\left \{ \widetilde{\mathbf{X}}_x^\text{train} =  \nabla_x \widetilde{\mathbf{S}}^\text{train}, \widetilde{\mathbf{X}}_y^\text{train} =  \nabla_y \widetilde{\mathbf{S}}^\text{train} \right \}$ and two-channel ground-truth mask will be $\left \{ \mathbf{v}_x^\text{train}, \mathbf{v}_y^\text{train} \right \}$ which corresponds to the sparse-codes $\Lambda = \left \{  i,j \in  \left \{  1,2,..,n_1\right \} \times \left \{  1,2,..,n_2\right \} : \left |\mathbf{\nabla} S_{i,j}  \right | > \tau_1   \right \}$. In the CS-aided recovery approach, when a test sample proxy is introduced to OSEN, the output probability map produced by the network is used as a likelihood measure of the corresponding support set. This is achieved as follows, given produced probability maps, $\left (\mathbf{p}_x, \mathbf{p}_y \right )$, we first compute the weights of the cost $\mathbf{\Gamma}_x = \frac{1}{\mathbf{p}_x + \epsilon} $ and $\mathbf{\Gamma}_y = \frac{1}{\mathbf{p}_y + \epsilon}$. Similar to \eqref{eq:weighted-lasso}, the weighted total-variation minimization problem can be expressed as follows:
\begin{equation}
    \min_{ \mathbf{S}} \left \{ \left \| \mathbf{y} -\mathbf{A} \text{vec}(\mathbf{S}) \right \|_2^2 + \lambda  \left \| \mathbf{\Gamma} \odot \mathbf{\nabla} \mathbf{S} \right \|_\text{TV} \right \},
    \label{eq:weighted_TV}
\end{equation}
where the second term is defined as,
\begin{equation}
\label{eq:weighted_TV_Gamma}
    \left \| \mathbf{\Gamma} \odot \mathbf{\nabla} \mathbf{S} \right \|_\text{TV}  =     \left \| \mathbf{\Gamma}_x \odot \mathbf{\nabla}_x\mathbf{S} \right \|_1 + \left \| \mathbf{\Gamma}_y \odot  \mathbf{\nabla}_y \mathbf{S} \right \|_1.
\end{equation}
The traditional ADMM solver can still be used in order to solve the proposed minimization scheme in \eqref{eq:weighted_TV} by modifying soft thresholding part of the solver to weighted soft thresholding. In this way, thanks to the proposed hybrid approach integrating model-based optimization procedure with the data-driven OSEN, we aim to improve the SR performance.

\section{Experimental Results}
\label{result}

The proposed approach is extensively evaluated in three different applications including the use-case scenarios where the signal/image to be sensed is sparse in spatial domain, representation-based classification, and CS/reconstruction of Magnetic Resonance Imaging (MRI). In this section, we will first give details about the experimental setup and we will present the results for different applications of the proposed OSEN approaches.

\subsection{Experimental Setup}

There are two compact configurations used in the proposed approach. The first configuration ($\text{OSEN}_1$) model has only two hidden operational layers with 48 and 24 neurons, respectively. The $\text{OSEN}_2$ architecture has max-pooling and transposed operational layers connected to the first hidden layer. Both networks have $3 \times 3$ kernel sizes and their activation functions are set to \textit{hyperbolic tangent} except for the output layers which are set to \textit{sigmoid} and \textit{softmax} for segmentation and classification, respectively.

The proposed approach is trained using Adam optimizer \cite{kingma2014adam} with its default learning parameters: learning rate $\alpha=10^{-3}$, $\beta_1 = 0.9$, and $\beta_2 = 0.999$ for 30 epochs in the representation-based classification and 100 epochs in the other applications. The regularization parameter $\lambda$ is searched empirically in log-scale over validation sets within the range of $\lambda^* \in [10^{-10}, 10^{2}]$, then it is fine-tuned with slight adjustments as $\lambda = \lambda^* \pm 10^{log(\lambda^*)}$. The proposed approach has been implemented using Python with the Tensorflow library \cite{abadi2016tensorflow} on a workstation having Intel ® i$9-7900$X CPU and NVidia ® 1080 Ti GPU with 128 GB system memory.

In the results, it is important to show performance improvements obtained by the proposed approach over the baseline model, that is the least-square sense solution with the CRC approach  \cite{collaborative}. The competing SRC approaches used in this study are Primal and Dual Augmented Lagrangian Methods, (PALM and DALM) \cite{fast}, $\ell_1$-regularized Least Squares ($\ell_1$-LS) \cite{l1ls}, ADMM \cite{ADMM}, Orthogonal Matching Pursuit (OMP) \cite{fast}, Homotopy \cite{homotopy}, Gradient Projection for Sparse Reconstruction (GPSR) \cite{gpsr}, and $\ell_1$-magic \cite{l1magic}. For a fair comparison, dictionary samples of the competing CRC and SRC approaches are enlarged by adding training samples used by the proposed approach. Since the SE literature is limited, in the comparisons, we modify several state-of-the-art SR approaches including ReconNet \cite{reconnet}, LAMP \cite{lamp}, LISTA \cite{lista}, and LVAMP \cite{lvamp} by training these reconstruction networks for the SE task using proposed input-output training pairs. Finally, we compare the proposed OSEN approach with our previous support estimator CSEN \cite{csen}. Above-mentioned deep learning approaches are implemented using the same Tensorflow environment following their proposed training configurations, whereas the SRC and CRC approaches have been tested with MATLAB version 2019a. For the learning-aided SR application, TV minimization via ADMM is performed on Python with the following parameter values: $\lambda = 0.01$, $\rho = 1.0$, $\alpha = 0.7$, $\text{abs\_tol}=10^{-4}$, $\text{rel\_tol}=10^{-2}$, $\text{max\_it}=2000$.

The results are reported using the following metrics: in representation-based classification, accuracy is computed as follows,
\begin{equation}
    \text{Accuracy} = (\text{TP} + \text{TN}) / (\text{TP} + \text{TN} + \text{FP} + \text{FN}),
\end{equation}
where the number of true positive and negative samples are TP and TN; false positive and negative samples are FP and FN, respectively. Next, SE performance is evaluated between the true binary mask $\mathbf{v}$ and estimated mask $\widehat{\mathbf{v}}$:
\begin{equation}
    \text{Precision} = \text{TP} / (\text{TP} + \text{FP}).
\end{equation}
\begin{equation}
    \text{Specificity} = \text{TN} / (\text{TN} + \text{FP}),
\end{equation}
\begin{equation}
    \text{Sensitivity} = \text{TP} / (\text{TP} + \text{FN}).
\end{equation}
Moreover, $F_1$ and $F_2$ scores are obtained by setting $\beta = 1$ and $\beta = 2$ in the following,
\begin{equation}
    F_\beta = (1 + \beta ^ 2) \frac{\text{Precision} \times \text{Sensitivity}}{\beta ^ 2 \times \text{Precision} + \text{Sensitivity}}.
\end{equation}
Therefore, $F_2$ score has more importance on improved TP detection than TN. After computing the metrics pixel-wise for each test sample, we report the averaged values using the Macro-average method. Finally, averaged PSNR and normalized MSE (NMSE) values are reported for learning-aided CS experiments. Note that we repeat all the reported experiments five times and present the final averaged values.

\begin{table*}[t!]
\caption{Support recovery performances are given for the proposed approach (bold) and competing methods using MNIST dateset. In CSEN configurations, ($+$) denotes that the number of neurons is increased two times in the hidden layers.}
\label{tab:mnist_results}
\centering
\setlength{\tabcolsep}{6pt} 
\renewcommand{\arraystretch}{0.85} 
\begin{tabular}{cccccccc}
\toprule
& \multicolumn{1}{c}{\textbf{Method}} & \multicolumn{1}{c}{\textbf{Precision ($\%$)}} & \multicolumn{1}{c}{\textbf{Specificity ($\%$)}} & \multicolumn{1}{c}{\textbf{Sensitivity ($\%$)}} & \multicolumn{1}{c}{\boldmath$F_1$\textbf{-Score} ($\%$)} & \multicolumn{1}{c}{\boldmath$F_2$\textbf{-Score} ($\%$)} & \multicolumn{1}{c}{\textbf{Accuracy} ($\%$)} \\ \midrule

\multicolumn{1}{c}{\multirow{26}{*}{\rotatebox[origin=c]{90}{\textbf{MR} $= $ \boldmath$5 \%$}}} & $\text{CSEN}_1$ \cite{csen} & $76.52 \pm 0.51$ & $94.15 \pm 0.21$ & $81.36 \pm 0.59$ & $78.86 \pm 0.42$ & $80.34 \pm 0.49$ & $91.56 \pm 0.22$ \\
                                                                          
& $\text{CSEN}_2$ \cite{csen} & $81.49 \pm 0.16$ & $95.43 \pm 0.07$ & $85.33 \pm 0.18$ & $83.37 \pm 0.15$ & $84.53 \pm 0.16$ & $93.39 \pm 0.08$ \\

& $\text{CSEN}_1+$ \cite{csen} & $79.34 \pm 0.44$ & $94.87 \pm 0.16$ & $83.74 \pm 0.22$ & $81.48 \pm 0.25$ & $82.82 \pm 0.19$ & $92.61 \pm 0.14$ \\

& $\text{CSEN}_2+$ \cite{csen} & $84.30 \pm 0.41$ & $96.15 \pm 0.13$ & $87.40 \pm 0.31$ & $85.82 \pm 0.30$ & $86.76 \pm 0.28$ & $94.39 \pm 0.14$ \\

& ReconNet \cite{reconnet} & $77.39 \pm 0.87$ & $94.23 \pm 0.30$ & $83.29 \pm 0.42$ & $80.23 \pm 0.51$ & $82.04 \pm 0.38$ & $91.95 \pm 0.24$ \\

& LAMP ($T = 2$) \cite{lamp} & $82.09 \pm 0.46$ & $94.69 \pm 0.10$ & $83.16 \pm 0.26$ & $82.62 \pm 0.15$ & $82.94 \pm 0.14$ & $92.36 \pm 0.14$ \\

& LAMP ($T = 3$) \cite{lamp} & $82.82 \pm 0.48$ & $94.12 \pm 0.31$ & $82.34 \pm 0.22$ & $82.58 \pm 0.25$ & $82.44 \pm 0.19$ & $91.74 \pm 0.24$ \\

& LAMP ($T = 4$) \cite{lamp} & $83.13 \pm 0.24$ & $94.23 \pm 0.16$ & $82.73 \pm 0.26$ & $82.93 \pm 0.15$ & $82.81 \pm 0.20$ & $91.90 \pm 0.12$ \\

& LISTA ($T = 2$) \cite{lista} & $82.84 \pm 0.43$ & $95.36 \pm 0.11$ & $82.98 \pm 0.17$ & $82.91 \pm 0.26$ & $82.95 \pm 0.19$ & $92.81 \pm 0.10$ \\

& LISTA ($T = 3$) \cite{lista} & $82.85 \pm 0.55$ & $95.36 \pm 0.15$ & $83.01 \pm 0.14$ & $82.93 \pm 0.26$ & $82.98 \pm 0.12$ & $92.82 \pm 0.11$ \\

& LISTA ($T = 4$) \cite{lista} & $82.85 \pm 0.58$ & $95.35 \pm 0.17$ & $83.13 \pm 0.21$ & $82.99 \pm 0.27$ & $83.08 \pm 0.16$ & $92.83 \pm 0.12$ \\

& LVAMP ($T = 1$) \cite{lvamp} & $73.49 \pm 1.03$ & $92.30 \pm 0.62$ & $74.75 \pm 0.60 $ & $74.11 \pm 0.48$ & $74.49 \pm 0.43$ & $89.10 \pm 0.39$ \\

& LVAMP ($T = 2$) \cite{lvamp} & $76.31 \pm 0.86$ & $88.32 \pm 1.31$ & $72.61 \pm 0.75$ & $74.41 \pm 0.55$ & $73.32 \pm 0.62$ & $85.40 \pm 1.11$ \\

& LVAMP ($T = 3$) \cite{lvamp} & $81.15 \pm 0.73$ & $94.52 \pm 0.27$ & $82.74 \pm 0.39$ & $81.94 \pm 0.50$ & $82.41 \pm 0.41$ & $92.08 \pm 0.27$ \\

& \textbf{OSEN}$_1$ ($Q = 1$) & $82.63 \pm 1.45$ & $95.71 \pm 0.39$ & $86.21 \pm 0.85$ & $84.38 \pm 1.16$ & $85.47 \pm 0.97$ & $93.80 \pm 0.49$ \\

& \textbf{OSEN}$_1$ ($Q = 3$) & $85.85 \pm 0.24$ & $96.55 \pm 0.06$ & $87.72 \pm 1.45$ & $86.77 \pm 0.81$ & $87.34 \pm 1.19$ & $94.80 \pm 0.29$ \\

& \textbf{OSEN}$_1$ ($Q = 5$) & $87.47 \pm 1.02$ & $97.02 \pm 0.33$ & $87.25 \pm 1.29$ & $87.34 \pm 0.27$ & $87.28 \pm 0.85$ & $95.09 \pm 0.09$ \\

& \textbf{OSEN}$_2$ ($Q = 1$) & $84.39 \pm 1.20$ & $96.17 \pm 0.28$ & $87.36 \pm 1.01$ & $85.85 \pm 1.10$ & $86.75 \pm 1.04$ & $94.41 \pm 0.44$ \\

& \textbf{OSEN}$_2$ ($Q = 3$) & $86.97 \pm 0.23$ & $96.80 \pm 0.06$ & $89.33 \pm 0.27$ & $88.13 \pm 0.17$ & $88.84 \pm 0.22$ & $95.32 \pm 0.06$ \\

& \textbf{OSEN}$_2$ ($Q = 5$) & $87.36 \pm 0.44$ & $96.91 \pm 0.13$ & $89.03 \pm 0.63$ & $88.19 \pm 0.16$ & $88.69 \pm 0.42$ & $95.36 \pm 0.05$ \\

& \textbf{NCL-OSEN}$_1$ ($Q = 1$) & $88.27 \pm 0.65$ & $97.13 \pm 0.17$ & $90.28 \pm 0.44$ & $89.26 \pm 0.54$ & $89.87 \pm 0.48$ & $95.78 \pm 0.22$ \\

& \textbf{NCL-OSEN}$_1$ ($Q = 3$) & $89.46 \pm 0.19$ & $97.42 \pm 0.07$ & $91.21 \pm 0.12$ & $90.33 \pm 0.11$ & $90.85 \pm 0.10$ & $96.20 \pm 0.04$ \\

& \textbf{NCL-OSEN}$_1$ ($Q = 5$) & $89.51 \pm 0.19$ & $97.42 \pm 0.04$ & $91.25 \pm 0.14$ & $90.37 \pm 0.11$ & $90.89 \pm 0.11$ & $96.21 \pm 0.04$ \\

& \textbf{NCL-OSEN}$_2$ ($Q = 1$) & $89.22 \pm 0.12$ & $97.37 \pm 0.03$ & $90.90 \pm 0.20$ & $90.05 \pm 0.13$ & $90.56 \pm 0.17$ & $96.09 \pm 0.05$ \\

& \textbf{NCL-OSEN}$_2$ ($Q = 3$) & $89.52 \pm 0.20$ & $\bm{97.43} \pm 0.05$ & $\bm{91.26} \pm 0.16$ & $90.38 \pm 0.11$ & $\bm{90.90} \pm 0.13$ & $\bm{96.22} \pm 0.04$ \\

& \textbf{NCL-OSEN}$_2$ ($Q = 5$) & $\bm{89.57} \pm 0.16$ & $\bm{97.43} \pm 0.06$ & $91.22 \pm 0.22$ & $\bm{90.39} \pm 0.13$ & $90.88 \pm 0.17 $  & $\bm{96.22} \pm 0.05$ \\ \midrule

\multicolumn{1}{c}{\multirow{28}{*}{\rotatebox[origin=c]{90}{\textbf{MR} $= $ \boldmath$10 \%$}}} & $\text{CSEN}_1$ \cite{csen} & $82.98 \pm 0.71$ & $95.95 \pm 0.19$ & $85.12 \pm 0.98$ & $84.04 \pm 0.79$ & $84.68 \pm 0.89$ & $93.73 \pm 0.32$ \\

& $\text{CSEN}_2$ \cite{csen} & $86.43 \pm 0.33$ & $96.75 \pm 0.08$ & $88.55 \pm 0.39$ & $87.47 \pm 0.35$ & $88.11 \pm 0.37$ & $95.09 \pm 0.14$ \\

& $\text{CSEN}_1+$ \cite{csen} & $85.0 \pm 0.60$ & $96.44 \pm 0.15$ & $87.10 \pm 0.79$ & $86.04 \pm 0.68$ & $86.67 \pm 0.74$ & $94.53 \pm 0.27$ \\

& $\text{CSEN}_2+$ \cite{csen} & $86.68 \pm 0.35$ & $97.30 \pm 0.10$ & $90.16 \pm 0.29$ & $89.41 \pm 0.21$ & $89.86 \pm 0.23$ & $95.87 \pm 0.09$ \\

& ReconNet \cite{reconnet} & $83.95 \pm 0.53$ & $96.12 \pm 0.14$ & $86.93 \pm 0.36$ & $85.41 \pm 0.41$ & $86.31 \pm 0.37$ & $94.21 \pm 0.16$ \\

& LAMP ($T = 2$) \cite{lamp} & $90.11 \pm 0.16$ & $96.75 \pm 0.13$ & $87.26 \pm 0.18$ & $88.66 \pm 0.07$ & $87.82 \pm 0.12$ & $94.92 \pm 0.08$ \\

& LAMP ($T = 3$) \cite{lamp} & $90.40 \pm 0.10$ & $96.57 \pm 0.09$ & $87.57 \pm 0.28$ & $88.96 \pm 0.11$ & $88.12 \pm 0.21$ & $94.83 \pm 0.07$ \\

& LAMP ($T = 4$) \cite{lamp} & $91.45 \pm 0.16$ & $97.54 \pm 0.07$ & $89.05 \pm 0.29$ & $90.23 \pm 0.10$ & $89.52 \pm 0.21$ & $95.90 \pm 0.04$ \\

& LISTA ($T = 2$) \cite{lista} & $90.21 \pm 0.28$ & $97.46 \pm 0.08$ & $87.03 \pm 0.07$ & $88.59 \pm 0.16$ & $87.64 \pm 0.10$ & $95.38 \pm 0.08$ \\

& LISTA ($T = 3$) \cite{lista} & $90.20 \pm 0.31$ & $97.45 \pm 0.09$ & $87.07 \pm 0.18$ & $88.61 \pm 0.16$ & $87.68 \pm 0.15$ & $95.38 \pm 0.08$ \\

& LISTA ($T = 4$) \cite{lista} & $90.32 \pm 0.31$ & $97.49 \pm 0.09$ & $86.98 \pm 0.18$ & $88.62 \pm 0.16$ & $87.63 \pm 0.15$ & $95.40 \pm 0.08$ \\

& LVAMP ($T = 1$) \cite{lvamp} & $79.46 \pm 0.85$ & $93.64 \pm 0.53$ & $78.90 \pm 0.64$ & $79.17 \pm 0.50$ & $79.01 \pm 0.51$ & $91.20 \pm 0.34$ \\

& LVAMP ($T = 2$) \cite{lvamp} & $82.68 \pm 0.41$ & $91.95 \pm 0.68$ & $74.22 \pm 0.42$ & $78.22 \pm 0.36$ & $75.77 \pm 0.39$ & $88.65 \pm 0.57$ \\

& LVAMP ($T = 3$) \cite{lvamp} & $87.91 \pm 0.55$ & $96.41 \pm 0.16$ & $86.06 \pm 0.21$ & $86.97 \pm 0.33$ & $86.42 \pm 0.23$ & $94.35 \pm 0.17$ \\

& \textbf{OSEN}$_1$ ($Q = 1$) & $87.96 \pm 0.66$ & $97.14 \pm 0.19$ & $88.97 \pm 0.66$ & $88.46 \pm 0.21$ & $88.76 \pm 0.43$ & $95.51 \pm 0.08$ \\

& \textbf{OSEN}$_1$ ($Q = 3$) & $89.81 \pm 0.55$ & $97.56 \pm 0.11$ & $90.81 \pm 0.77$ & $90.31 \pm 0.65$ & $90.61 \pm 0.72$ & $96.23 \pm 0.25$ \\

& \textbf{OSEN}$_1$ ($Q = 5$) & $90.94 \pm 0.73$ & $97.86 \pm 0.21$ & $90.47 \pm 0.62$ & $90.70 \pm 0.11$ & $90.56 \pm 0.36$ & $96.41 \pm 0.06$ \\

& \textbf{OSEN}$_2$ ($Q = 1$) & $88.44 \pm 1.21$ & $97.26 \pm 0.27$ & $89.16 \pm 1.62$ & $88.80 \pm 1.34$ & $89.01 \pm 1.50$ & $95.65 \pm 0.52$ \\

& \textbf{OSEN}$_2$ ($Q = 3$) & $90.87 \pm 0.32$ & $97.81 \pm 0.08$ & $91.74 \pm 0.19$ & $91.30 \pm 0.13$ & $91.56 \pm 0.12$ & $96.62 \pm 0.06$ \\

& \textbf{OSEN}$_2$ ($Q = 5$) & $91.13 \pm 0.61$ & $97.90 \pm 0.16$ & $91.33 \pm 0.64$ & $91.23 \pm 0.31$ & $91.29 \pm 0.46$ & $96.60 \pm 0.12$ \\

& \textbf{NCL-OSEN}$_1$ ($Q = 1$) & $93.39 \pm 0.21$ & $98.42 \pm 0.06$ & $93.35 \pm 0.23$ & $93.37 \pm 0.04$ & $93.36 \pm 0.15$ & $97.45 \pm 0.01$ \\

& \textbf{NCL-OSEN}$_1$ ($Q = 3$) & $93.49 \pm 0.04$ & $98.43 \pm 0.01$ & $93.94 \pm 0.07$ & $93.72 \pm 0.02$ & $93.85 \pm 0.05$ & $97.57 \pm 0.01$ \\

& \textbf{NCL-OSEN}$_1$ ($Q = 5$) & $\bm{93.55} \pm 0.07$ & $\bm{98.44} \pm 0.02$ & $93.94 \pm 0.08$ & $\bm{93.74} \pm 0.02$ & $\bm{93.86} \pm 0.05$ & $\bm{97.58} \pm 0.01$ \\

& \textbf{NCL-OSEN}$_2$ ($Q = 1$) & $93.32 \pm 0.07$ & $98.39 \pm 0.03$ & $93.84 \pm 0.13$ & $93.58 \pm 0.05$ & $93.74 \pm 0.09$ & $97.52 \pm 0.02$ \\

& \textbf{NCL-OSEN}$_2$ ($Q = 3$) & $93.46 \pm 0.13$ & $98.42 \pm 0.04$ & $\bm{93.96} \pm0.13 $ & $93.71 \pm 0.05$ & $\bm{93.86} \pm 0.09$ & $97.57 \pm 0.02$ \\

& \textbf{NCL-OSEN}$_2$ ($Q = 5$) & $93.49 \pm 0.17$ & $98.43 \pm 0.05$ & $93.88 \pm 0.17$ & $93.68 \pm 0.01$ & $93.80 \pm 0.10$ & $97.56 \pm 0.01$ \\ \midrule

\multicolumn{1}{c}{\multirow{28}{*}{\rotatebox[origin=c]{90}{\textbf{MR} $= $ \boldmath$25 \%$}}} & $\text{CSEN}_1$ \cite{csen} & $89.26 \pm 0.16$ & $97.53 \pm 0.04$ & $90.03 \pm 0.27$ & $89.65 \pm 0.21$ & $89.88 \pm 0.24$ & $95.96 \pm 0.09$ \\

& $\text{CSEN}_2$ \cite{csen} & $91.12 \pm 0.26$ & $97.93 \pm 0.07$ & $91.71 \pm 0.26$ & $91.41 \pm 0.21$ & $91.59 \pm 0.23$ & $96.66 \pm 0.09$ \\

& $\text{CSEN}_1+$ \cite{csen} & $90.37 \pm 0.19$ & $97.78 \pm 0.05$ & $91.04 \pm 0.25$ & $90.71 \pm 0.18$ & $90.91 \pm 0.21$ & $96.38 \pm 0.08$ \\

& $\text{CSEN}_2+$ \cite{csen} & $92.17 \pm 0.16$ & $98.16 \pm 0.04$ & $92.93 \pm 0.18$ & $92.55 \pm 0.17$ & $92.78 \pm 0.17$ & $97.11 \pm 0.07$ \\

& ReconNet \cite{reconnet} & $88.57 \pm 0.37$ & $97.36 \pm 0.09$ & $89.52 \pm 0.21$ & $89.04 \pm 0.24$ & $89.33 \pm 0.20$ & $95.72 \pm 0.11$ \\

& LAMP ($T = 2$) \cite{lamp} & $93.76 \pm 0.64$ & $96.05 \pm 1.01$ & $88.28 \pm 0.93$ & $90.93 \pm 0.74$ & $89.32 \pm 0.85$ & $94.60 \pm 0.99$ \\

& LAMP ($T = 3$) \cite{lamp} & $95.04 \pm 0.15$ & $98.43 \pm 0.10$ & $91.51 \pm 0.21$ & $93.24 \pm 0.07$ & $92.20 \pm 0.15$ & $97.33 \pm 0.08$ \\

& LAMP ($T = 4$) \cite{lamp} & $94.98 \pm 0.11$ & $98.39 \pm 0.05$ & $91.87 \pm 0.10$ & $93.40 \pm 0.06$ & $92.48 \pm 0.08$ & $97.40 \pm 0.03$ \\

& LISTA ($T = 2$) \cite{lista} & $95.25 \pm 0.08$ & $98.71 \pm 0.04$ & $90.13 \pm 0.10$ & $92.62 \pm 0.04$ & $91.11 \pm 0.07$ & $97.09 \pm 0.03$ \\

& LISTA ($T = 3$) \cite{lista} & $95.35 \pm 0.11$ & $98.74 \pm 0.04$ & $90.03 \pm 0.07$ & $92.62 \pm 0.03$ & $91.05 \pm 0.04$ & $97.10 \pm 0.03$ \\

& LISTA ($T = 4$) \cite{lista} & $95.40 \pm 0.20$ & $98.75 \pm 0.65$ & $90.01 \pm 0.15$ & $92.63 \pm 0.02$ & $91.04 \pm 0.09$ & $97.10 \pm 0.03$ \\

& LVAMP ($T = 1$) \cite{lvamp} & $81.44 \pm 0.77$ & $93.88 \pm 0.32$ & $81.20 \pm 0.91$ & $81.31 \pm 0.30$ & $81.24 \pm 0.62$ & $91.91 \pm 0.15$ \\

& LVAMP ($T = 2$) \cite{lvamp} & $89.57 \pm 0.30$ & $96.19 \pm 0.19$ & $75.80 \pm 0.34$ & $82.11 \pm 0.27$ & $78.21 \pm 0.31$ & $92.40 \pm 0.20$ \\

& LVAMP ($T = 3$) \cite{lvamp} & $94.93 \pm 0.15$ & $98.38 \pm 0.09$ & $89.84 \pm 0.20$ & $92.32 \pm 0.08$ & $90.82 \pm 0.15$ & $96.80 \pm 0.09$ \\

& \textbf{OSEN}$_1$ ($Q = 1$) & $90.81 \pm 0.66$ & $97.84 \pm 0.15$ & $91.45 \pm 0.53$ & $91.13 \pm 0.58$ & $91.32 \pm 0.55$ & $96.56 \pm 0.22$ \\

& \textbf{OSEN}$_1$ ($Q = 3$) & $92.54 \pm 0.76$ & $98.24 \pm 0.19$ & $92.69 \pm 0.43$ & $92.61 \pm 0.58$ & $92.66 \pm 0.48$ & $97.15 \pm 0.23$ \\

& \textbf{OSEN}$_1$ ($Q = 5$) & $93.02 \pm 0.22$ & $98.34 \pm 0.06$ & $93.30 \pm 0.27$ & $93.16 \pm 0.11$ & $93.25 \pm 0.19$ & $97.36 \pm 0.05$ \\

& \textbf{OSEN}$_2$ ($Q = 1$) & $91.59 \pm 0.71$ & $98.01 \pm 0.17$ & $92.33 \pm 0.64$ & $91.96 \pm 0.67$ & $92.18 \pm 0.65$ & $96.89 \pm 0.27$ \\

& \textbf{OSEN}$_2$ ($Q = 3$) & $93.29 \pm 0.21$ & $98.40 \pm 0.06$ & $93.77 \pm 0.11$ & $93.53 \pm 0.11$ & $93.67 \pm 0.10$ & $97.50 \pm 0.05$ \\

& \textbf{OSEN}$_2$ ($Q = 5$) & $93.33 \pm 0.58$ & $98.43 \pm 0.14$ & $93.31 \pm 0.59$ & $93.32 \pm 0.47$ & $93.32 \pm 0.52$ & $97.42 \pm 0.18$ \\

& \textbf{NCL-OSEN}$_1$ ($Q = 1$) & $\bm{95.89} \pm 0.14$ & $\bm{99.02} \pm 0.04$ & $95.65 \pm 0.18$ & $95.77 \pm 0.02$ & $95.70 \pm 0.12$ & $98.37 \pm 0.01$ \\

& \textbf{NCL-OSEN}$_1$ ($Q = 3$) & $95.83 \pm 0.13$ & $99.0 \pm 0.04$ & $95.78 \pm 0.15$ & $95.80 \pm 0.02$ & $95.79 \pm 0.10$ & $\bm{98.38} \pm 0.01$ \\

& \textbf{NCL-OSEN}$_1$ ($Q = 5$) & $95.81 \pm 0.05$ & $98.98 \pm 0.02$ & $\bm{95.82} \pm 0.05$ & $\bm{95.81} \pm 0.01$ & $\bm{95.82} \pm 0.03$ & $\bm{98.38} \pm 0.01$ \\

& \textbf{NCL-OSEN}$_2$ ($Q = 1$) & $95.68 \pm 0.11$ & $98.96 \pm 0.02$ & $95.80 \pm 0.08$ & $95.74 \pm 0.02$ & $95.78 \pm 0.04$ & $98.36 \pm 0.01$ \\

& \textbf{NCL-OSEN}$_2$ ($Q = 3$) & $95.79 \pm 0.06$ & $98.99 \pm 0.02$ & $95.69 \pm 0.03$ & $95.74 \pm 0.03$ & $95.71 \pm 0.03$ & $98.36 \pm 0.01$ \\

& \textbf{NCL-OSEN}$_2$ ($Q = 5$) & $95.68 \pm 0.06$ & $98.96 \pm 0.02$ & $95.79 \pm 0.09$ & $95.74 \pm 0.02$ & $95.77 \pm 0.06$ & $98.36 \pm 0.01 $ \\ \bottomrule

\end{tabular}
\end{table*}

\subsection{Support Estimation from CS measurements}

The proposed approach is first evaluated over the MNIST dataset with handwritten digit samples to show its ability to perform support recovery when the signal itself is already sparse in spatial domain. The dataset has $70 000$ samples and it is divided into train, validation, and test splits with the ratio of $5:1:1$, respectively. Each sample image in the dataset has a size of $28 \times 28$ with the intensity range in $[0, 1]$. A sample from the MNIST dataset is an example of a sparse signal since the foreground is more predominant compared to the digits. In fact, its averaged sparsity ratio is computed as $\rho = \frac{k}{n} = ~ 0.2$ for the vectorized samples $\mathbf{x} \in \mathbb{R}^{784}$. Since the samples are sparse in spatial domain or canonical basis, the sparsifying domain is set to $\mathbf{\Phi} =\mathbf{I}$ and \eqref{Eq:CS} can be written as,
\begin{equation}
    \mathbf{y} = \mathbf{A} \mathbf{x} = \mathbf{D} \mathbf{x},
\end{equation}
with $\mathbf{A} = \mathbf{D} \in \mathbb{R}^{m \times n}$. The elements of the measurement matrix, $A_{i,j}$, are i.i.d. drawn from $\mathcal{N}\left ( 0,\frac{1}{m} \right )$.

\begin{figure*}[h]
     \centering
     \begin{subfigure}[b]{0.32\linewidth}
         \centering
         \includegraphics[width=\linewidth]{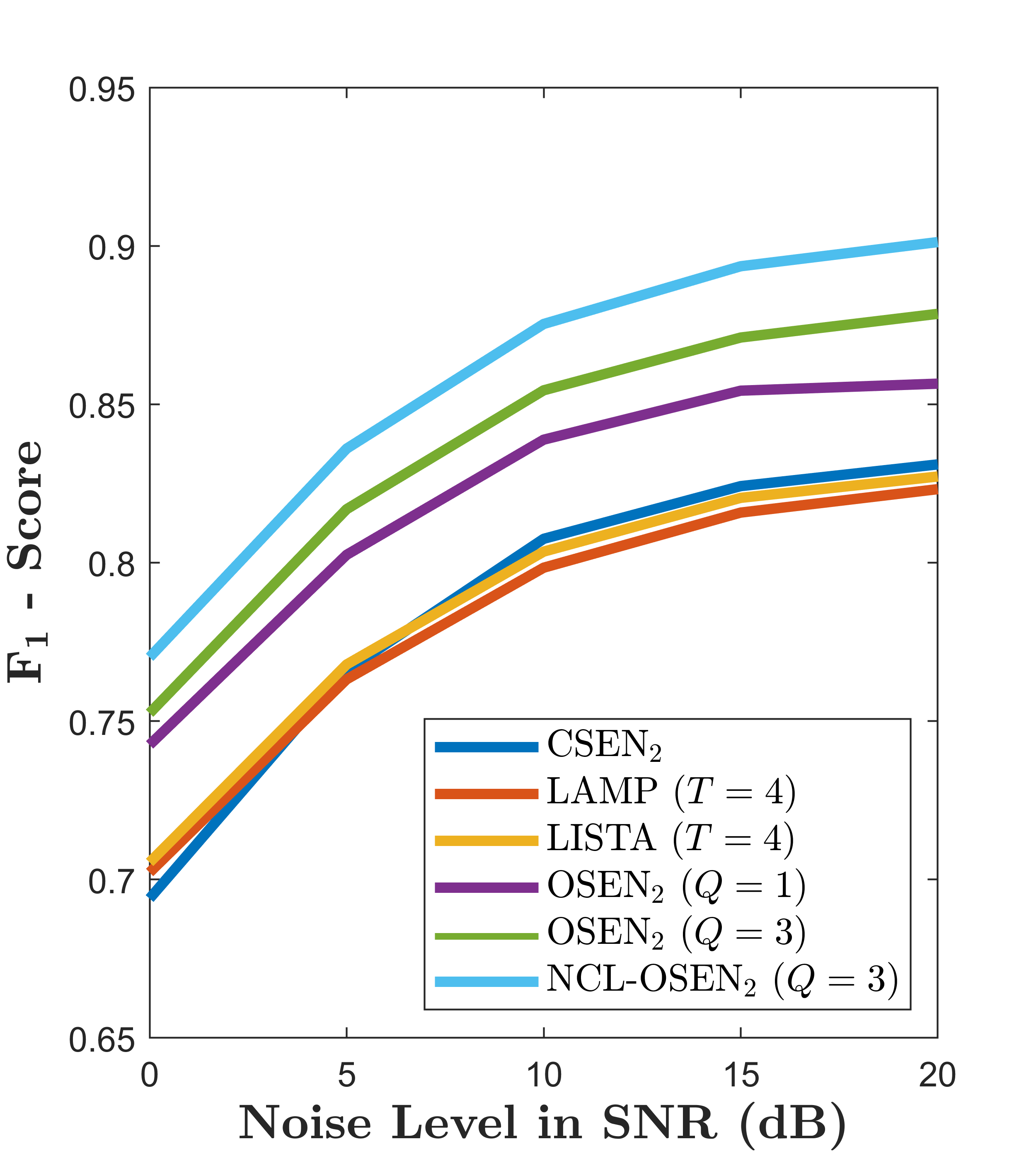}
         \caption{$\text{MR}$ $=$ $0.05$ $(m/n)$}
     \end{subfigure}
     \hfill
     \begin{subfigure}[b]{0.32\linewidth}
         \centering
         \includegraphics[width=\linewidth]{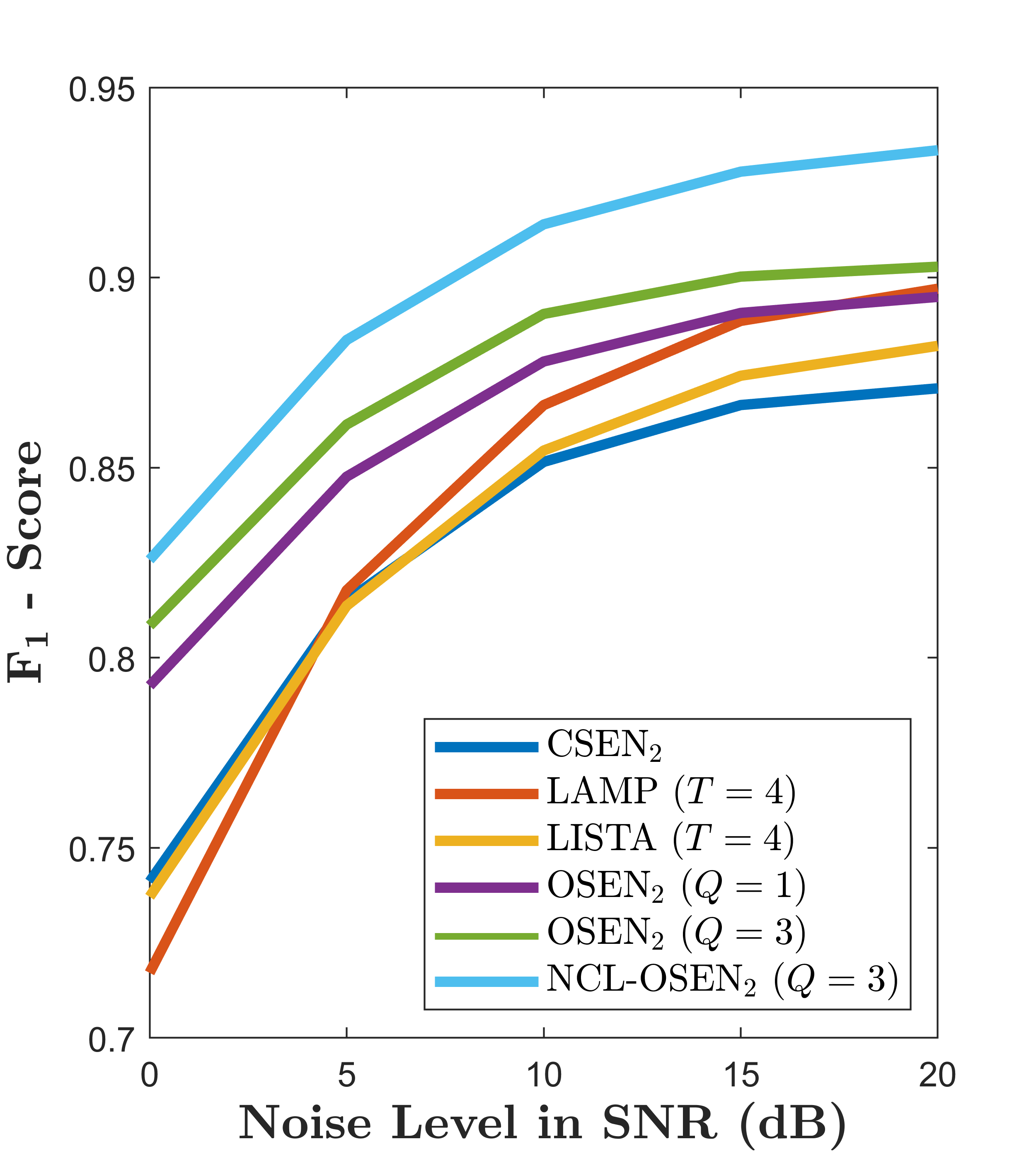}
         \caption{$\text{MR}$ $=$ $0.1$ $(m/n)$}
     \end{subfigure}
     \hfill
     \begin{subfigure}[b]{0.32\linewidth}
         \centering
         \includegraphics[width=\linewidth]{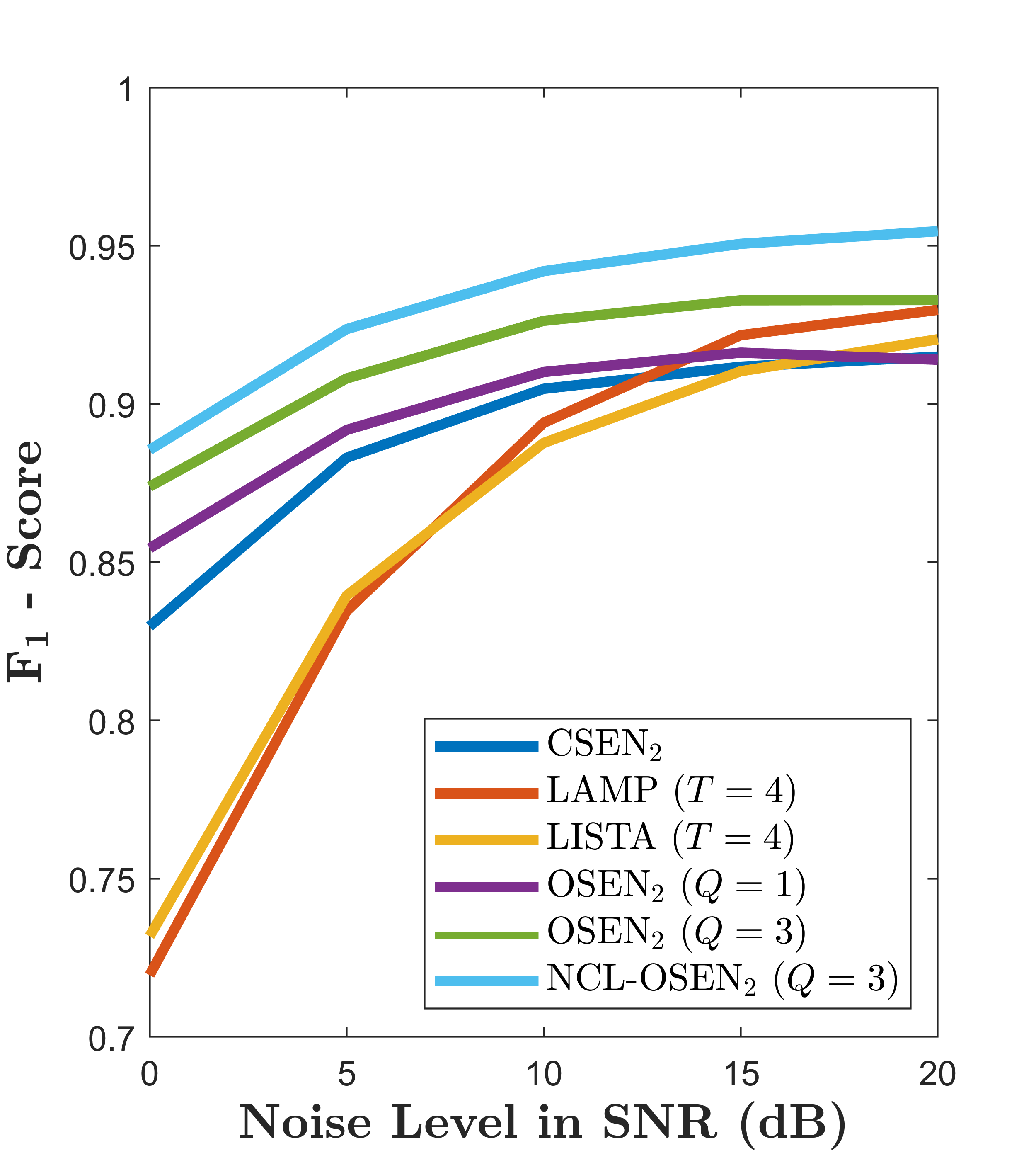}
         \caption{$\text{MR}$ $=$ $0.25$ $(m/n)$}
     \end{subfigure}
        \caption{Obtained $F_1$-Scores by the methods under noisy measurements using different Measurement Rates (MRs).}
        \label{fig:noisy_mnist}
\end{figure*}

In the MNIST experiments, for a measurement test sample $\mathbf{y}$, the input proxy is computed as $\widetilde{\mathbf{x}} = \mathbf{B}\mathbf{y}$ where $\mathbf{B} = \mathbf{D}^T$. Then, the reshaped proxy with size of $28 \times 28$ is given as input to the OSENs. In the results, we change MR from $0.05$ to $0.25$ to observe its effect on the performance. The results are presented in Table \ref{tab:mnist_results} for the competing methods and proposed approach using different configurations. For a better comparison, the number of linear transformation layers ($T$) has been varied also for LAMP, LISTA, and LVAMP methods. It is seen that the proposed OSEN approaches outperform other approaches with significant performance margins in $F_1$ and $F_2$-scores. For example, the gap in $F_1$-scores reaches approximately $8\%$ and $5\%$ comparing $\text{OSEN}_1$ with $\text{CSEN}_1$ and $\text{OSEN}_2$ with $\text{CSEN}_2$, respectively, when the polynomial order is set to $Q = 3$. Moreover, the proposed NCL-OSEN framework equipped with the NCL scheme further increases the estimation performance achieving the state-of-the-art $F_1$-score larger than the previous competitor $\text{CSEN}_1$ and $\text{CSEN}_2$ approaches approximately by $11.5\%$ and $7\%$, respectively. The performance improvement of using higher $Q$ values for the NCL method is limited for $\text{OSEN}_1$ and has no noticeable effect on $\text{OSEN}_2$ configuration when MR is $0.25$. For higher $Q$ values, it is possible to learn superior kernel mapping functions with higher-order Taylor approximation which can be beneficial for complex learning problems. In general, the SE task becomes less challenging for higher MR values and most of the competing methods can achieve fair performance levels, whereas small MR settings can benchmark the performance of the methods. One can say that further improvement on the SE performance is limited considering $F_1$ and $F_2$-scores are already greater than $95\%$. In Table \ref{tab:mnist_results}, even if we increase the number of neurons for CSENs, they are still outperformed by the proposed approach. Note the fact that LAMP method is designed for Gaussian measurement matrix, but it still lacks the performance which has been achieved by the proposed approach. The number of trainable parameters and elapsed times are given in Table \ref{tab:mnist_comp} when MR is set to $0.05$. It is shown that OSENs have computational efficiency compared to other approaches. We increase the number of parameters in CSEN$+$ configuration to have approximately similar computational complexity with the OSENs. As shown in Table \ref{tab:mnist_results}, CSEN$+$ still cannot reach the same performance level with the OSENs.

The robustness of the methods to measurement noise is evaluated by introducing Gaussian additive noise to the measurement, $\mathbf{y} = \mathbf{D}\mathbf{x} + \mathbf{z}$. Achieved $F_1$-Scores are presented for varied noise levels in Fig. \ref{fig:noisy_mnist}. Accordingly, it has been observed that the proposed approach still achieves \textit{state-of-the-art} performance levels even under severe noisy conditions. Among the competing methods, the CSEN approach shows some robustness to the noise. Even though, especially for low MR values, it has a similar decaying trend, the proposed OSEN and NCL-OSEN approaches have an overall robustness to the measurement noise for all MR values compared to the competing methods. This is not surprising considering the superior learning capabilities of the super-neurons and generative perceptrons.

\begin{table}[]
\centering
\caption{Number of trainable parameters and elapsed times per sample using the aforementioned PC setup are given on MNIST dataset for $\text{MR} = 0.05$. CSEN$+$ has two times more neurons in the hidden layers.}
\label{tab:mnist_comp}
 \begin{tabular}{@{}ccc@{}}
\toprule
\textbf{Method}          & \textbf{Parameters (\#)} & \textbf{Time (sec)} \\ \midrule
CSEN$_1$ \cite{csen}           & 11,089               & 0.15     \\
CSEN$_2$ \cite{csen}           & 16,297               & 0.15     \\
CSEN$_1$+ \cite{csen}          & 42,913               & 0.15     \\
CSEN$_2$+ \cite{csen}          & 63,697               & 0.16     \\
ReconNet \cite{reconnet}        & 72,643               & 0.19     \\
LAMP ($T=2$) \cite{lamp}      & 645,234              & 5.37     \\
LAMP ($T=3$) \cite{lamp}      & 645,235              & 5.59     \\
LAMP ($T=4$) \cite{lamp}      & 645,236              & 7.3      \\
LISTA ($T=2$) \cite{lista}     & 30,582               & 5.49     \\
LISTA ($T=3$) \cite{lista}     & 30,585               & 5.29     \\
LISTA ($T=4$) \cite{lista}     & 30,588               & 5.47     \\
LVAMP ($T=1$) \cite{lvamp}     & 32,928               & 5.91     \\
LVAMP ($T=2$) \cite{lvamp}     & 680,512              & 7.42     \\
LVAMP ($T=3$) \cite{lvamp}     & 1,328,096            & 7.65     \\
OSEN$_1$ ($Q=1$)     & 11,235               & 0.29     \\
OSEN$_1$ ($Q=3$)     & 33,413               & 0.31     \\
OSEN$_1$ ($Q=5$)     & 55,591               & 0.34     \\
OSEN$_2$ ($Q=1$)     & 16,491               & 0.36     \\
OSEN$_2$ ($Q=3$)     & 49,085               & 0.41     \\
OSEN$_2$ ($Q=5$)     & 81,679               & 0.44     \\
NCL-OSEN$_1$ ($Q=1$) & 42,595               & 0.29     \\
NCL-OSEN$_1$ ($Q=3$) & 127,493              & 0.32     \\
NCL-OSEN$_1$ ($Q=5$) & 212,391              & 0.35     \\
NCL-OSEN$_2$ ($Q=1$) & 47,851               & 0.35     \\
NCL-OSEN$_2$ ($Q=3$) & 143,165              & 0.4      \\
NCL-OSEN$_2$ ($Q=5$) & 238,479              & 0.43     \\ \bottomrule
\end{tabular}
\end{table}

\begin{figure}[!h]
\centering
\vspace{-0.005cm}
\subfloat[$\text{MR}$ $=$ $0.01$ $(m/n)$]{%
  \includegraphics[clip,width=0.9\columnwidth]{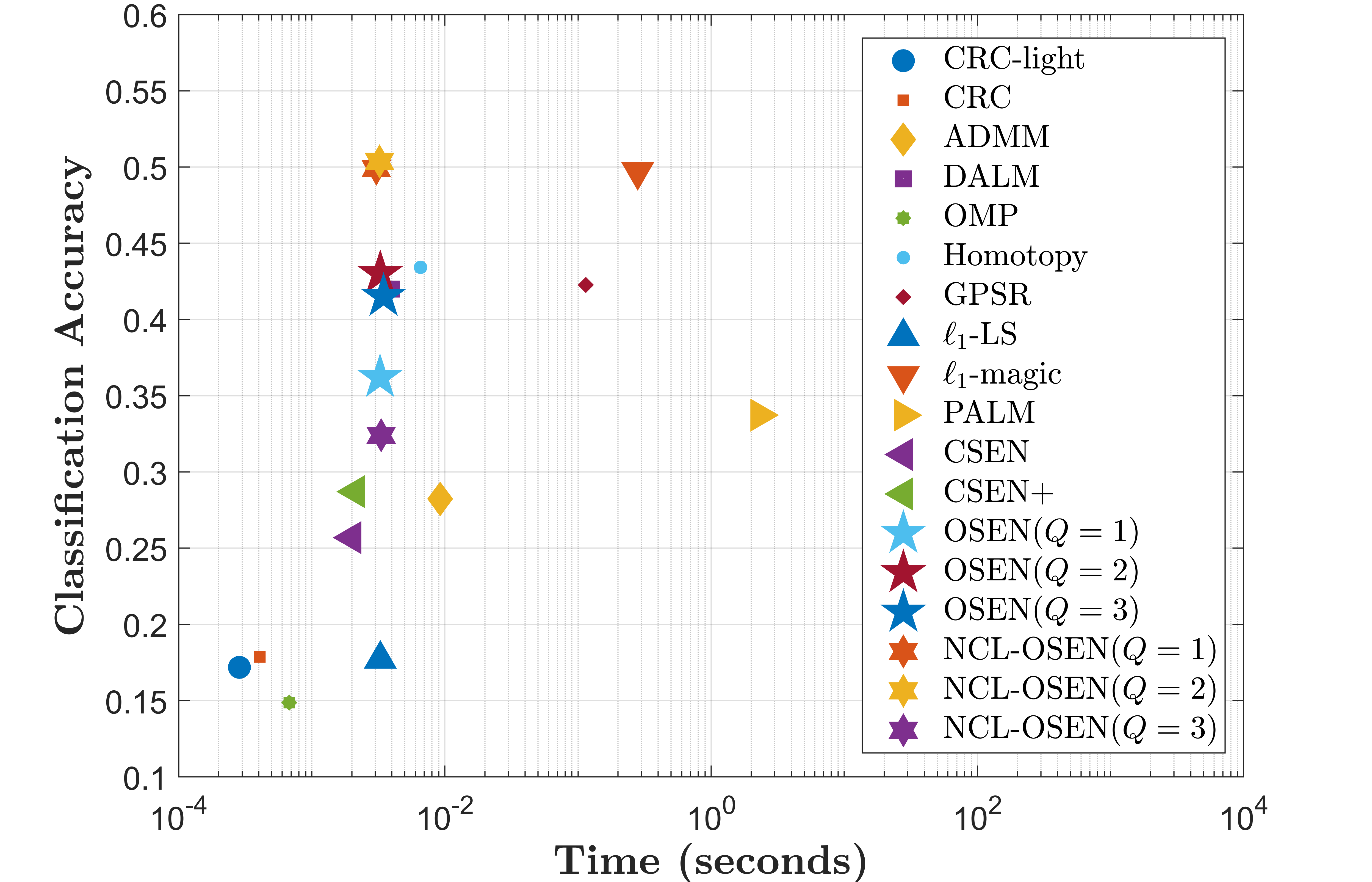}%
}
\vspace{-0.005cm}
\subfloat[$\text{MR}$ $=$ $0.05$ $(m/n)$]{%
  \includegraphics[clip,width=0.9\columnwidth]{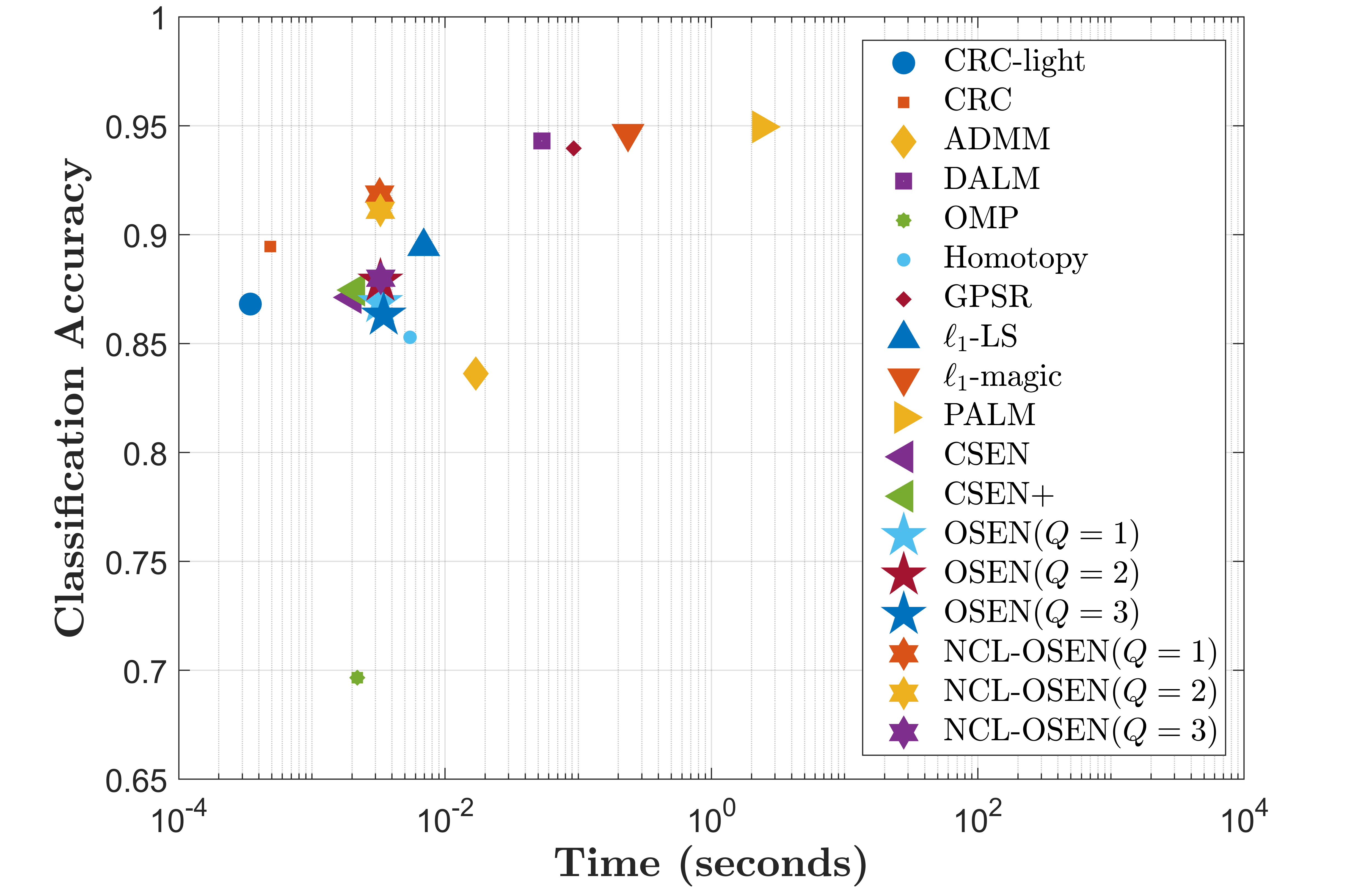}%
}
\vspace{-0.005cm}
\subfloat[$\text{MR}$ $=$ $0.25$ $(m/n)$]{%
  \includegraphics[clip,width=0.9\columnwidth]{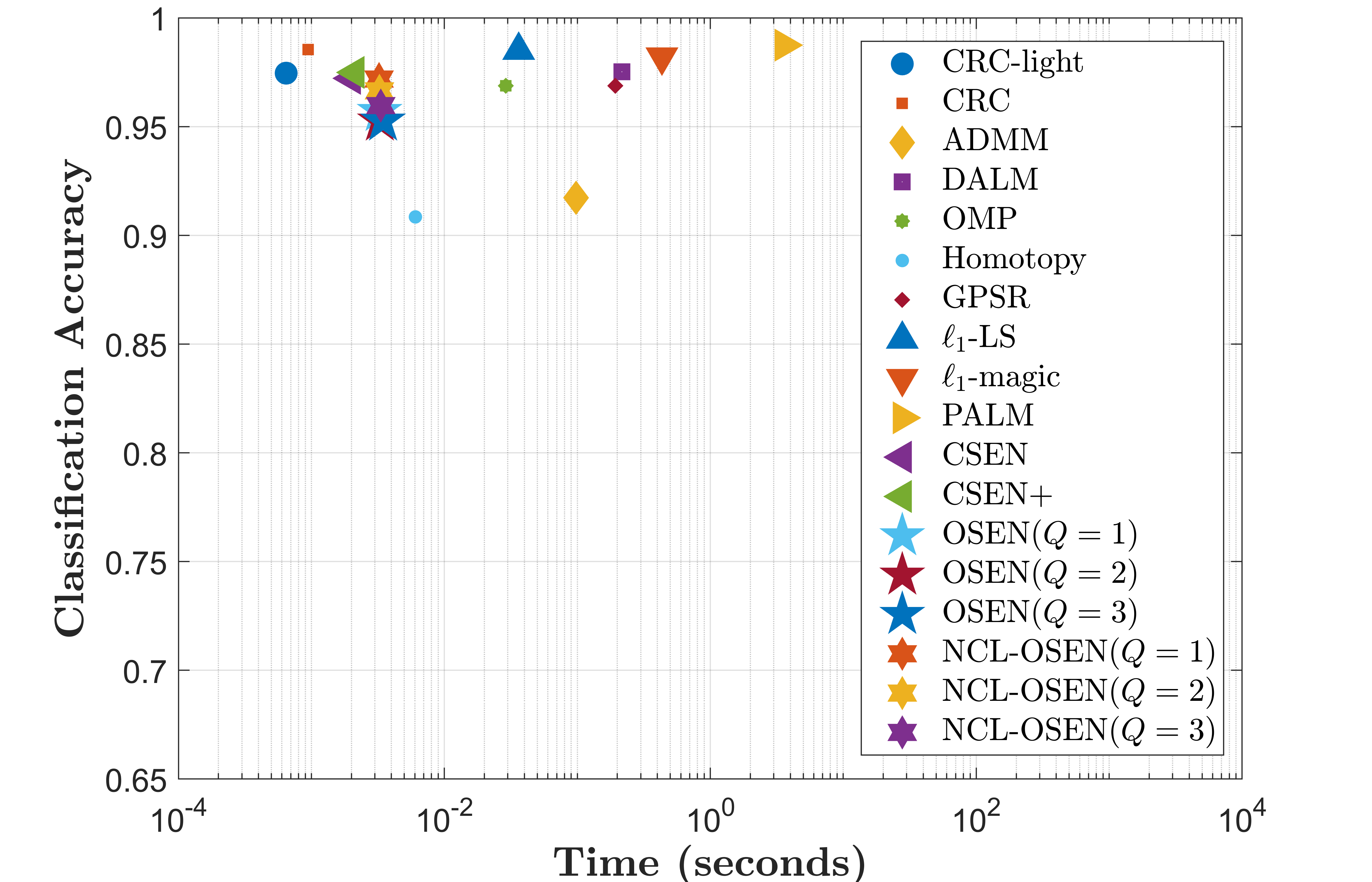}%
}
\caption{Classification accuracy versus computational time for the proposed OSEN and competing approaches. CSEN$+$ has two times more neurons in the hidden layers.}
\label{fig:yale_results}
\end{figure}

\subsection{Face Recognition using Representation-based Classification}

Face recognition can be considered as a representation-based classification problem and it is well-suited to evaluate the proposed approach because of the limited number of available samples from each identity. We use Yale-B \cite{yale} dataset consisting of $2414$ samples with $32 \times 32$ pixel sizes belonging to $38$ different identities. Note that since all experiments are repeated five times, splits for dictionary, train, validation, and test are different in each run. To evaluate the competing CRC and SRC methods, $48$ and $16$ samples per identity are randomly selected to for the dictionary and test set, respectively. The learning-based methods, OSEN and CSEN, use $25\%$ of data from the dictionary samples during training from which $25\%$ of the samples are separated for the validation set. Consequently, their dictionaries are constructed using only $32$ samples per identity for a fair comparison against CRC and SRC approaches. More specifically, the collected dictionary with vectorized samples would have a size of $1024 \times 1216$. Then, we use PCA matrix $\mathbf{A}$ for dimensionality reduction to obtain the equivalent dictionary $\mathbf{D}$ in \eqref{Eq:CS}. In face recognition experiments, for a given query sample $\mathbf{y}$, coarse estimation is computed using the denoiser matrix as $\widetilde{\mathbf{x}} = \mathbf{B}\mathbf{y}$ where $\mathbf{B} = \left ( \mathbf{D}^T \mathbf{D} + \lambda \mathbf{I} \right )^{-1} \mathbf{D}^T$. Next, the input proxy $\widetilde{\mathbf{x}} \in \mathbb{R}^{1216}$ is reshaped to obtain a 2-D proxy image with the size of $16 \times 76$. This reshaping is performed by following a specific ordering to make sure that the dictionary samples belonging the same class categories are grouped together. Hence, the kernel size of the average-pooling operation is set to the size of grouped pixels from the same class (i.e., $8 \times 4$).

Achieved classification accuracies are presented in Fig. \ref{fig:yale_results} along with required computational times in the log-scale. We only report accuracies obtained by the initial network structures, $\text{OSEN}_1$ and $\text{CSEN}_1$, since no significant performance improvements have been observed by using transposed operational and convolutional layers in the second network configuration. As shown in Fig. \ref{fig:yale_results},  CSENs have similar computational time with the proposed approach; and hence, we consider the CSEN approach as previous \textit{state-of-the-art} since we aim for a maximum classification accuracy with a minimum computational time complexity. The CRC-light approach has the same number of samples in the dictionary as the proposed approach. It is observed that especially for low MRs, the proposed approach significantly improves the classification accuracy. For instance, the accuracy gap is around $25\%$ between the NCL-OSENs and CSENs when $\text{MR} = 0.01$. On the other hand, when MR is large enough, the classification problem becomes less complicated and all methods can achieve higher than $90\%$ accuracy. Overall, the proposed approach is able to achieve comparable or better performance levels than computationally complex and iterative SRC methods. This demonstrates that the proposed solution is highly robust across all measurement rates but especially for lower rates, it has a significant superiority. On the other hand, the performance of different SR algorithms is inconsistent and varies significantly from one setup to another \cite{csen}.  Note also the fact that contrary to the previous results in Table \ref{tab:mnist_results}, small values for the polynomial order $Q$ are chosen in Yale-B dataset.

\begin{figure*}[!h]
    \centering
    \includegraphics[width=0.95\linewidth]{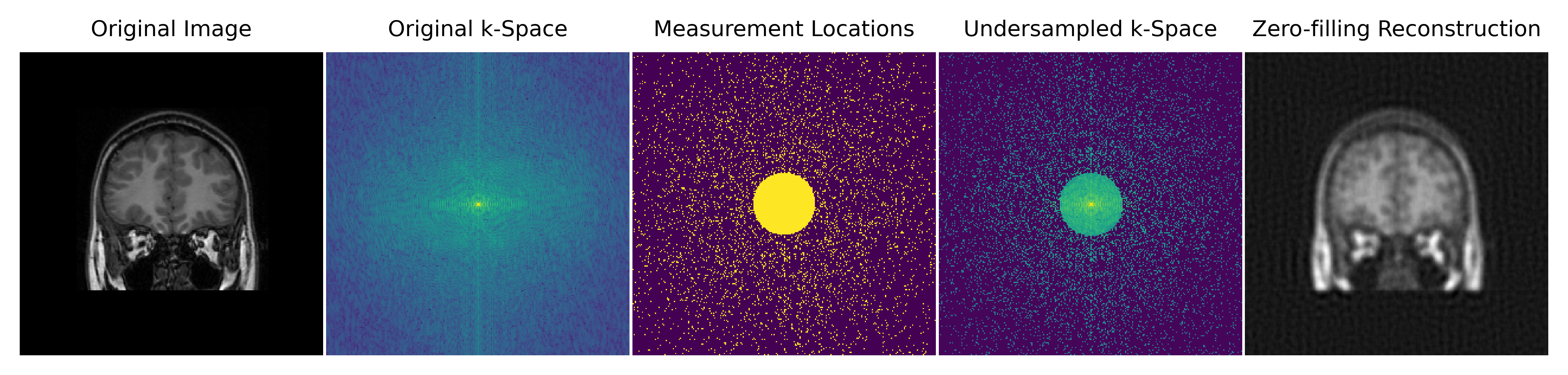}
    \caption{Semi-random sampling is applied for an MRI image with Gaussian and $\ell_2$-ball to sample discrete Fourier measurements, then the zero-filling reconstruction is obtained by applying inverse DFT using selected samples.}     
\label{fig:sampling}
\end{figure*}

\begin{table*}[h]
\vspace{0.1cm}
\caption{MRI reconstruction performance of the proposed approaches with different Q settings are compared against the reconstruction by zero-filling, baseline method with TV-minimization, and two CSEN approaches. The highest PSNR values achieved are in bold.}
\label{tab:tv_results}
\centering
\resizebox{0.8\linewidth}{!}{\begin{tabular}{ccccccc}
\toprule
\multicolumn{1}{c}{} & \multicolumn{3}{c}{\textbf{PSNR (dB)}} & \multicolumn{3}{c}{\textbf{NMSE ($\%$)}} \\ \cmidrule(lr){2-4} \cmidrule(lr){5-7}

\multicolumn{1}{c}{\textbf{MR}} & \multicolumn{1}{c}{\textbf{0.05}} & \multicolumn{1}{c}{\textbf{0.1}} & \multicolumn{1}{c}{\textbf{0.25}} & \multicolumn{1}{c}{\textbf{0.05}} & \multicolumn{1}{c}{\textbf{0.1}} & \multicolumn{1}{c}{\textbf{0.25}} \\ \midrule

Zero-filling & $27.72 \pm 0.04$ & $30.66 \pm 0.04$ & $36.56 \pm 0.05$ & $6.20 \pm 0.04$ & $3.21 \pm 0.02$ & $0.85 \pm 0.01$ \\

TV (Baseline) \cite{tv2} & $29.07 \pm 0.04$ & $33.27 \pm 0.04$ & $43.23 \pm 0.05$ & $4.60 \pm 0.04$ & $1.85 \pm 0.02$ & $0.23 \pm 0.01$ \\

CSEN$_1$ \cite{csen} & $29.90 \pm 0.07$ & $34.71 \pm 0.05$ & $44.36 \pm 0.02$ & $3.80 \pm 0.06$ & $1.35 \pm 0.01$ & $0.15 \pm 0.01$ \\

CSEN$_2$ \cite{csen} & $30.05 \pm 0.07$ & $34.89 \pm 0.03$ & $44.51 \pm 0.05$ & $3.67 \pm 0.05$ & $1.29 \pm 0.01$ & $0.15 \pm 0.01$ \\

\textbf{OSEN}$_1$ ($Q = 1$) & $30.19 \pm 0.08$ & $34.96 \pm 0.04$ & $44.64 \pm 0.05$ & $3.58 \pm 0.06$ & $1.28 \pm 0.01$ & $0.15 \pm 0.01$ \\

\textbf{OSEN}$_1$ ($Q = 2$) & $30.33 \pm 0.09$ & $35.14 \pm 0.04$ & $44.83 \pm 0.04$ & $3.45 \pm 0.07$ & $1.22 \pm 0.01$ & $0.14 \pm 0.01$ \\

\textbf{OSEN}$_1$ ($Q = 3$) & $30.32 \pm 0.06$ & $35.12 \pm 0.04$ & $44.85 \pm 0.06$ & $3.46 \pm 0.04$ & $1.23 \pm 0.01$ & $0.14 \pm 0.01$ \\

\textbf{OSEN}$_2$ ($Q = 1$) & $30.38 \pm 0.07$ & $35.16 \pm 0.05$ & $44.90 \pm 0.05$ & $3.41 \pm 0.05$ & $1.21 \pm 0.01$ & $\bm{0.13} \pm 0.01$ \\

\textbf{OSEN}$_2$ ($Q = 2$) & $30.48 \pm 0.07$ & $35.27 \pm 0.05$ & $44.90 \pm 0.04$ & $3.33 \pm 0.05$ & $\bm{1.18} \pm 0.01$ & $\bm{0.13} \pm 0.01$ \\

\textbf{OSEN}$_2$ ($Q = 3$) & $\bm{30.49} \pm 0.06$ & $35.26 \pm 0.04$ & $44.90 \pm 0.06$ & $\bm{3.32} \pm 0.04$ & $\bm{1.18} \pm 0.01$ & $\bm{0.13} \pm 0.01$ \\ 

\textbf{NCL-OSEN}$_1$ ($Q = 1$) & $30.44 \pm 0.08$ & $35.11 \pm 0.06$ & $45.11 \pm 0.07$ & $3.37 \pm 0.06$ & $1.24 \pm 0.02$ & $\bm{0.13} \pm 0.01$ \\

\textbf{NCL-OSEN}$_1$ ($Q = 2$) & $30.48 \pm 0.06$ & $35.22 \pm 0.07$ & $45.07 \pm 0.09$ & $3.34 \pm 0.06$ & $1.20 \pm 0.02$ & $\bm{0.13} \pm 0.01$ \\

\textbf{NCL-OSEN}$_1$ ($Q = 3$) & $30.48 \pm 0.09$ & $35.28 \pm 0.03$ & $45.19 \pm 0.05$ & $3.33 \pm 0.06$ & $1.19 \pm 0.01$ & $\bm{0.13} \pm 0.01$ \\

\textbf{NCL-OSEN}$_2$ ($Q = 1$) & $30.45 \pm 0.08$ & $35.21 \pm 0.05$ & $45.02 \pm 0.09$ & $3.36 \pm 0.06$ & $1.21 \pm 0.01$ & $\bm{0.13} \pm 0.01$ \\

\textbf{NCL-OSEN}$_2$ ($Q = 2$) & $30.46 \pm 0.08$ & $35.24 \pm 0.05$ & $45.07 \pm 0.07$ & $3.35 \pm 0.05$ & $1.19 \pm 0.01$ & $\bm{0.13} \pm 0.01$ \\

\textbf{NCL-OSEN}$_2$ ($Q = 3$) & $\bm{30.49} \pm 0.06$ & $\bm{35.32} \pm 0.04$ & $\bm{45.20} \pm 0.08$ & $3.33 \pm 0.04$ & $\bm{1.18} \pm 0.01$ & $\bm{0.13} \pm 0.01$ \\\bottomrule
\end{tabular}}
\end{table*}

\subsection{Learning-aided Signal Reconstruction (SR) for Compressively Sensed (CS) MRI}

As the third application of the proposed approach, we consider CS MRI reconstruction using learning-aided SR. Assuming an MRI image to be reconstructed is $\mathbf{S} \in \mathbb{R}^{N \times N}$, the sensing model is defined as $\mathbf{y} = \mathbf{A} \mathcal{F} \text{vec}(\mathbf{S})$ where $\mathcal{F}$ is DFT matrix and $\mathbf{A} \in \mathbb{C}^{m \times n}$ is the sampling matrix to select spectral indices from the DFT. Choosing the sparsifying domain as $\mathbf{\Phi} = \mathbf{\nabla}$, the proposed weighted-TV minimization in \eqref{eq:weighted_TV} can be modified for the CS MRI reconstruction as follows,
\begin{equation}
    \min_{ \mathbf{S}} \left \{ \left \| \mathbf{y} -\mathbf{A} \mathcal{F} \text{vec}(\mathbf{S}) \right \|_2^2 + \lambda  \left \| \mathbf{\Gamma} \odot \mathbf{\nabla} \mathbf{S} \right \|_\text{TV} \right \},
    \label{eq:mri_weighted_TV}
\end{equation}
where $\mathbf{y} \in \mathbb{C}^{m}$ is the measurement. Computational complexity for computing $\mathbf{A} \mathcal{F} \text{vec}(\mathbf{S})$ and $\mathbf{A}^T \mathbf{y}$ increases significantly when the image size is large. Therefore, it is more feasible to use a structural measurement matrix for sampling indices of the Fourier domain and it is also the hardware requirements of the current MRI devices. The structural measurement matrix is constructed by semi-random sampling \cite{cs_book} using Gaussian and $\ell_2$ ball. More specifically, let a measurement set be $\mathbf{\Omega} = \mathbf{\Omega_1} \cup \mathbf{\Omega_2} \subseteq \{-n/4, ..., n/4\}^2$ with measurements $m_1 + m_2 = m$. Accordingly, the subset $\mathbf{\Omega}_1$ contains $m_1$ number of frequencies whose indices are drawn from standard normal distribution on $\{-n/4, ..., n/4 \}^2\ \setminus \mathbf{\Omega}_2$. The second set of frequencies are $\mathbf{\Omega}_2 = \{\omega_{i, j}\}$ where $i^2 + j^2 < r^2$ with $r = \sqrt{\frac{m/3}{\pi}}$ to make sure that a third of measurements are from the $\ell_2$-ball, since $\pi r^2 = m/3$. This proposed semi-random sampling scheme is illustrated in Fig. \ref{fig:sampling}.

\begin{figure*}[h]
    \centering
    \includegraphics[width=1\linewidth]{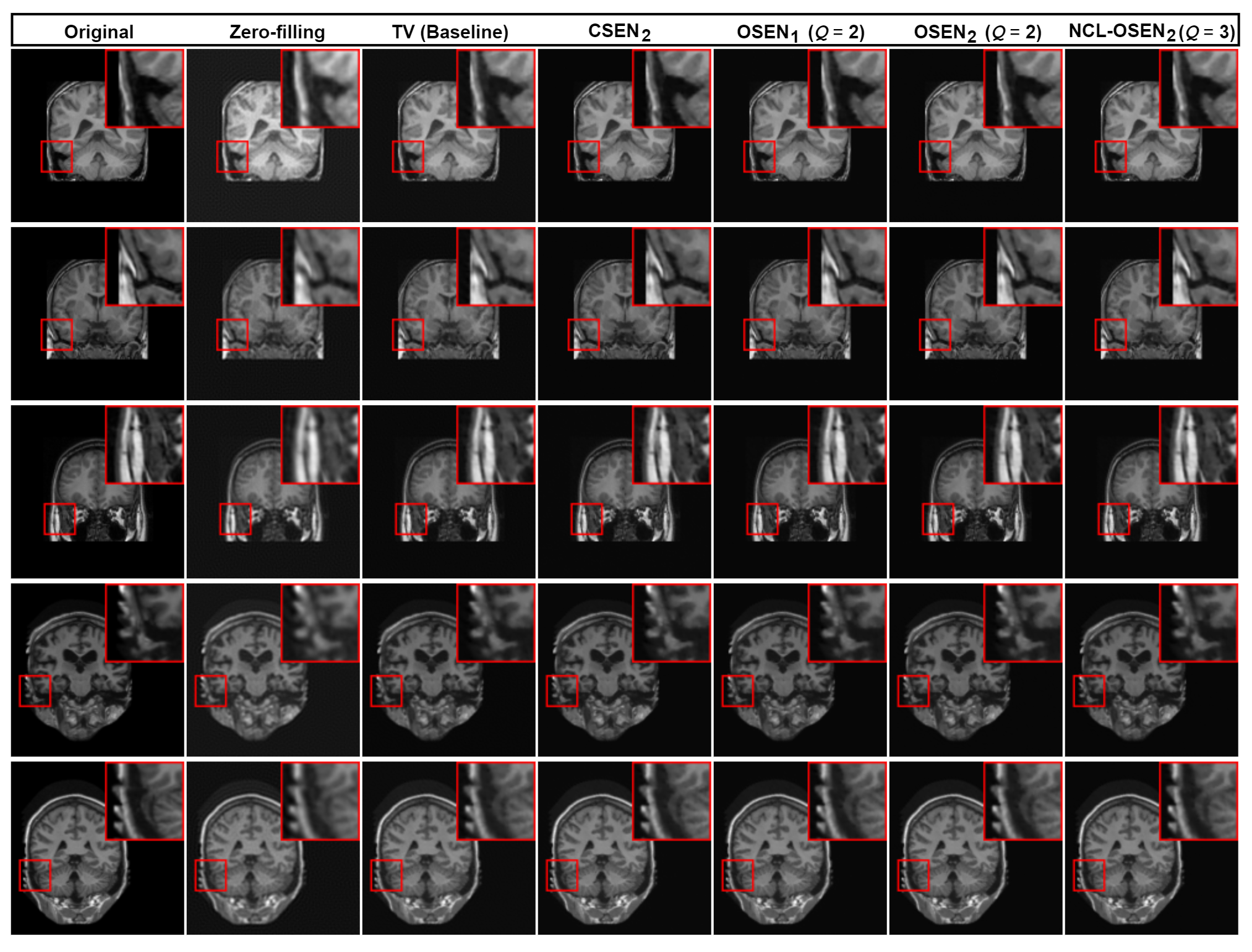}
    \vspace{-0.6cm}
    \caption{Recovered MRI images are presented using TV-based reconstruction technique with the proposed learning-aided SR approach for $\text{MR} = 0.25$. }
    \vspace{-0.5cm}
\label{fig:tv_results}
\end{figure*}

\begin{table*}[htbp]
\caption{Support recovery performances are presented for the proposed approaches with different Q settings, and the competing methods in the gradient domain for the MRI dataset.}
\vspace{-0.1cm}
\label{tab:mri_support_results}
\centering
\setlength{\tabcolsep}{6pt} 
\renewcommand{\arraystretch}{0.8} 
\begin{tabular}{cccccccc}
\toprule
& \multicolumn{1}{c}{\textbf{Method}} & \multicolumn{1}{c}{\textbf{Precision ($\%$)}} & \multicolumn{1}{c}{\textbf{Specificity ($\%$)}} & \multicolumn{1}{c}{\textbf{Sensitivity ($\%$)}} & \multicolumn{1}{c}{\boldmath$F_1$\textbf{-Score} ($\%$)} & \multicolumn{1}{c}{\boldmath$F_2$\textbf{-Score} ($\%$)} & \multicolumn{1}{c}{\textbf{Accuracy} ($\%$)} \\ \midrule

\multicolumn{1}{c}{\multirow{14}{*}{\rotatebox[origin=c]{90}{\textbf{MR} $= $ \boldmath$5 \%$}}} & $\text{CSEN}_1$ \cite{csen} & $58.54 \pm 4.12$ & $96.69 \pm 0.90$ & $47.63 \pm 4.99$ & $52.13 \pm 1.35$ & $49.26 \pm 3.68$ & $92.43 \pm 0.39$ \\
                                       
& $\text{CSEN}_2$ \cite{csen} & $\bm{68.83} \pm 2.66$ & $\bm{98.29} \pm 0.34$ & $39.13 \pm 2.86$ & $49.75 \pm 1.59$ & $42.77 \pm 2.52$ & $\bm{93.15} \pm 0.07$ \\

& \textbf{OSEN}$_1$ ($Q = 1$) & $51.12 \pm 0.77$ & $94.90 \pm 0.21$ & $57.12 \pm 0.65$ & $53.94 \pm 0.20$ & $55.80 \pm 0.34$ & $91.52 \pm 0.14$ \\

& \textbf{OSEN}$_1$ ($Q = 2$) & $52.78 \pm 1.34$ & $94.87 \pm 0.29$ & $60.13 \pm 0.32$ & $56.21 \pm 0.63$ & $58.50 \pm 0.16$ & $91.85 \pm 0.24$ \\

& \textbf{OSEN}$_1$ ($Q = 3$) & $53.17 \pm 1.30$ & $94.96 \pm 0.30$ & $60.03 \pm 0.51$ & $56.38 \pm 0.55$ & $58.51 \pm 0.20$ & $91.93 \pm 0.23$ \\

& \textbf{OSEN}$_2$ ($Q = 1$) & $51.99 \pm 2.30$ & $94.46 \pm 0.61$ & $62.69 \pm 1.18$ & $56.79 \pm 0.89$ & $60.17 \pm 0.28$ & $91.70 \pm 0.45$ \\

& \textbf{OSEN}$_2$ ($Q = 2$) & $53.45 \pm 2.89$ & $94.55 \pm 0.82$ & $65.18 \pm 1.37$ & $58.65 \pm 1.33$ & $62.37 \pm 0.22$ & $92.0 \pm 0.63$ \\

& \textbf{OSEN}$_2$ ($Q = 3$) & $51.80 \pm 1.63$ & $94.15 \pm 0.42$ & $65.84 \pm 0.52$ & $57.96 \pm 0.84$ & $62.44 \pm 0.22$ & $91.69 \pm 0.34$ \\

& \textbf{NCL-OSEN}$_1$ ($Q = 1$) & $59.11 \pm 1.09$ & $91.26 \pm 0.26$ & $70.77 \pm 1.58$ & $64.42 \pm 1.25$ & $68.09 \pm 1.42$ & $88.16 \pm 0.40$ \\

& \textbf{NCL-OSEN}$_1$ ($Q = 2$) & $60.86 \pm 2.96$ & $91.47 \pm 1.49$ & $73.41 \pm 4.73$ & $\bm{66.35} \pm 1.28$ & $70.36 \pm 3.02$ & $88.73 \pm 0.69$ \\

& \textbf{NCL-OSEN}$_1$ ($Q = 3$) & $61.68 \pm 0.83$ & $92.10 \pm 0.35$ & $71.17 \pm 1.80$ & $66.07 \pm 0.91$ & $69.03 \pm 1.37$ & $88.93 \pm 0.26$ \\

& \textbf{NCL-OSEN}$_2$ ($Q = 1$) & $59.79 \pm 1.42$ &  $91.35 \pm 0.58$ &  $71.97 \pm 2.04$ &  $65.30 \pm 1.12$ &  $69.14 \pm 1.52$ &  $88.41 \pm 0.44$ \\

& \textbf{NCL-OSEN}$_2$ ($Q = 2$) &  $59.01 \pm 1.75$ &  $91.03 \pm 0.66$ &  $72.15 \pm 1.55$ &  $64.90 \pm 1.25$ &  $69.06 \pm 1.25$ & $88.17 \pm 0.55$ \\

& \textbf{NCL-OSEN}$_2$ ($Q = 3$) & $59.88 \pm 1.91$ & $91.15 \pm 0.64$ & $\bm{73.83} \pm 1.31$ & $66.12 \pm 1.50$ & $\bm{70.53} \pm 1.31$ & $88.53 \pm 0.64$ \\ \midrule

\multicolumn{1}{c}{\multirow{14}{*}{\rotatebox[origin=c]{90}{\textbf{MR} $= $ \boldmath$10 \%$}}} & $\text{CSEN}_1$ \cite{csen} & $\bm{78.90} \pm 5.89$ & $\bm{98.80} \pm 0.56$ & $43.10 \pm 8.52$ & $54.78 \pm 6.50$ & $47.04 \pm 8.07$ & $93.96 \pm 0.26$ \\

& $\text{CSEN}_2$ \cite{csen} & $76.49 \pm 8.26$ & $98.36 \pm 1.05$ & $48.34 \pm 10.60$ & $57.74 \pm 5.27$ & $51.51 \pm 8.91$ & $\bm{94.02} \pm 0.14$ \\

& \textbf{OSEN}$_1$ ($Q = 1$) & $59.50 \pm 1.39$ & $95.88 \pm 0.26$ & $63.52 \pm 0.25$ & $61.43 \pm 0.66$ & $62.66 \pm 0.20$ & $93.07 \pm 0.22$ \\

& \textbf{OSEN}$_1$ ($Q = 2$) & $60.17 \pm 1.47$ & $95.74 \pm 0.29$ & $67.53 \pm 0.49$ & $63.62 \pm 0.64$ & $65.90 \pm 0.20$ & $93.29 \pm 0.22$ \\

& \textbf{OSEN}$_1$ ($Q = 3$) & $60.62 \pm 1.51$ & $95.83 \pm 0.29$ & $67.27 \pm 0.29$ & $63.76 \pm 0.72$ & $65.82 \pm 0.17$ & $93.35 \pm 0.24$ \\

& \textbf{OSEN}$_2$ ($Q = 1$) & $59.72 \pm 1.86$ & $95.57 \pm 0.37$ & $68.81 \pm 0.35$ & $63.92 \pm 0.93$ & $66.76 \pm 0.24$ & $93.25 \pm 0.30$ \\

& \textbf{OSEN}$_2$ ($Q = 2$) & $65.21 \pm 6.62$ & $96.53 \pm 1.27$ & $64.54 \pm 7.25$ & $64.13 \pm 0.89$ & $64.21 \pm 4.70$ & $93.75 \pm 0.54$ \\

& \textbf{OSEN}$_2$ ($Q = 3$) & $63.39 \pm 5.42$ & $96.16 \pm 1.02$ & $67.44 \pm 6.22$ & $64.83 \pm 0.88$ & $66.27 \pm 4.13$ & $93.66 \pm 0.39$ \\

& \textbf{NCL-OSEN}$_1$ ($Q = 1$) & $61.78 \pm 2.56$ & $92.50 \pm 0.81$ & $67.86 \pm 2.44$ & $64.64 \pm 2.07$ & $66.52 \pm 2.15$ & $88.75 \pm 0.78$ \\

& \textbf{NCL-OSEN}$_1$ ($Q = 2$) & $64.03 \pm 2.10$ & $92.91 \pm 0.63$ & $70.57 \pm 3.14$ & $67.11 \pm 2.08$ & $69.13 \pm 2.60$ & $89.53 \pm 0.63$ \\

& \textbf{NCL-OSEN}$_1$ ($Q = 3$) & $66.37 \pm 1.23$ & $93.54 \pm 0.55$ & $71.28 \pm 2.31$ & $68.70 \pm 0.61$ & $70.22 \pm 1.57$ & $90.17 \pm 0.18$ \\

& \textbf{NCL-OSEN}$_2$ ($Q = 1$) & $64.03 \pm 2.10$ &  $92.91 \pm 0.63$ &  $70.57 \pm 3.14$ &  $67.11 \pm 2.08$ &  $69.13 \pm 2.60$ &  $89.53 \pm 0.63$ \\

& \textbf{NCL-OSEN}$_2$ ($Q = 2$) &  $62.86 \pm 3.44$ &  $92.35 \pm 1.30$ &  $71.80 \pm 2.04$ &  $66.95 \pm 1.89$ &  $69.75 \pm 1.48$ & $89.24 \pm 0.98$ \\

& \textbf{NCL-OSEN}$_2$ ($Q = 3$) & $65.65 \pm 0.84$ & $93.18 \pm 0.22$ & $\bm{73.02} \pm 1.17$ & $\bm{69.14} \pm 0.86$ & $\bm{71.42} \pm 1.02$ & $90.12 \pm 0.27$ \\ \midrule

\multicolumn{1}{c}{\multirow{14}{*}{\rotatebox[origin=c]{90}{\textbf{MR} $= $ \boldmath$25 \%$}}} & $\text{CSEN}_1$ \cite{csen} & $\bm{92.29} \pm 4.19$ & $\bm{99.52} \pm 0.35$ & $52.73 \pm 9.57$ & $66.32 \pm 6.42$ & $57.35 \pm 8.71$ & $95.46 \pm 0.52$ \\

& $\text{CSEN}_2$ \cite{csen} & $92.12 \pm 6.23$ & $99.46 \pm 0.49$ & $42.83 \pm 26.07$ & $51.98 \pm 28.09$ & $45.91 \pm 26.93$ & $94.54 \pm 1.87$ \\

& \textbf{OSEN}$_1$ ($Q = 1$) & $76.57 \pm 3.53$ & $97.86 \pm 0.51$ & $72.11 \pm 3.92$ & $74.09 \pm 0.69$ & $72.85 \pm 2.64$ & $95.62 \pm 0.15$ \\

& \textbf{OSEN}$_1$ ($Q = 2$) & $81.44 \pm 0.48$ & $98.44 \pm 0.06$ & $72.02 \pm 0.62$ & $76.44 \pm 0.15$ & $73.72 \pm 0.45$ & $96.14 \pm 0.01$ \\

& \textbf{OSEN}$_1$ ($Q = 3$) & $79.71 \pm 2.61$ & $98.18 \pm 0.37$ & $74.37 \pm 2.99$ & $76.84 \pm 0.38$ & $75.31 \pm 1.97$ & $96.11 \pm 0.08$ \\

& \textbf{OSEN}$_2$ ($Q = 1$) & $80.84 \pm 0.34$ & $98.37 \pm 0.04$ & $72.04 \pm 0.35$ & $76.18 \pm 0.08$ & $73.64 \pm 0.24$ & $96.09 \pm 0.01$ \\

& \textbf{OSEN}$_2$ ($Q = 2$) & $81.26 \pm 0.51$ & $98.40 \pm 0.07$ & $72.69 \pm 0.80$ & $76.73 \pm 0.27$ & $74.26 \pm 0.59$ & $96.17 \pm 0.03$ \\

& \textbf{OSEN}$_2$ ($Q = 3$) & $80.98 \pm 2.34$ & $98.34 \pm 0.33$ & $73.63 \pm 2.56$ & $77.06 \pm 0.27$ & $74.95 \pm 1.66$ & $\bm{96.19} \pm 0.08$ \\

& \textbf{NCL-OSEN}$_1$ ($Q = 1$) & $70.67 \pm 1.91$ & $89.95 \pm 1.05$ & $\bm{89.97} \pm 1.54$ & $79.12 \pm 0.57$ & $\bm{85.28} \pm 0.59$ & $89.95 \pm 0.50$ \\

& \textbf{NCL-OSEN}$_1$ ($Q = 2$) & $74.37 \pm 2.91$ & $91.98 \pm 1.32$ & $86.20 \pm 0.86$ & $79.81 \pm 1.57$ & $83.51 \pm 0.71$ & $90.76 \pm 0.95$ \\

& \textbf{NCL-OSEN}$_1$ ($Q = 3$) & $75.38 \pm 1.25$ & $92.34 \pm 0.58$ & $87.28 \pm 1.02$ & $\bm{80.88} \pm 0.41$ & $84.60 \pm 0.53$ & $91.27 \pm 0.29$ \\

& \textbf{NCL-OSEN}$_2$ ($Q = 1$) & $73.63 \pm 1.11$ &  $91.77 \pm 0.48$ &  $85.66 \pm 1.04$ &  $79.18 \pm 0.80$ &  $82.94 \pm 0.85$ &  $90.48 \pm 0.41$ \\

& \textbf{NCL-OSEN}$_2$ ($Q = 2$) &  $74.77 \pm 2.14$ &  $92.32 \pm 0.93$ &  $84.50 \pm 1.38$ &  $79.31 \pm 0.98$ &  $82.33 \pm 0.86$ & $90.67 \pm 0.59$ \\

& \textbf{NCL-OSEN}$_2$ ($Q = 3$) & $74.19 \pm 0.82$ & $91.90 \pm 0.44$ & $86.73 \pm 1.94$ & $79.96 \pm 0.79$ & $83.88 \pm 1.39$ & $90.81 \pm 0.30$ \\ \bottomrule
\end{tabular}
\end{table*}

When the measurement domain is complex, we adapt the following changes in the NCL module. First, the NCL module operates in 2D where under-sampled frequency representation is obtained by $\mathbf{y}_f = \mathbf{A}\text{vec}(\mathbf{F})$ using sampling matrix $\mathbf{A}$ applied after 2D-DFT, i.e., $\mathbf{F} = \mathcal{F} \mathbf{S} \mathcal{F}^T$, where $\mathcal{F}$ is the Fourier matrix. Then, we apply zero-filling and reshaping to $\mathbf{y}_f$ in order to obtain $\mathbf{Y} \in \mathbb{C}^{N \times N}$ as input to the NCL module. Accordingly, 2D-Self-GOP transformation in the complex domain can be written as,
\begin{equation}
    \label{eq:complex_perceptron}
    \widetilde{\mathbf{X}} = \phi(\mathbf{Y}) = \sigma \left( \sum_{q=1}^Q \mathbf{W}_q \mathbf{Y}^{\odot q} \mathbf{W}^T_q + \mathbf{B}_q \right),
\end{equation}
where $\mathbf{W}_q \in \mathbb{C}^{N \times N}$ and $\mathbf{B}_q \in \mathbb{C}^{N \times N}$. The complex domain mapping in Self-GOPs is implemented as follows, for example, consider $\mathbf{b} = \mathbf{W}\mathbf{Y} = \left(\Re \left( \mathbf{W} \right) + i\Im \left(\mathbf{W} \right) \right) \left(\Re \left(\mathbf{Y}\right) + i\Im \left(\mathbf{Y} \right) \right)$, then the real and imaginary parts of the multiplication can be separately computed by $\Re\left( \mathbf{b} \right) = \Re \left( \mathbf{W} \right) \Re \left( \mathbf{Y} \right) + \Im \left( \mathbf{W} \right) \Im \left(\mathbf{Y}\right)$ and $\Im \left(\mathbf{b} \right) = \Im \left(\mathbf{W} \right) \Re \left(\mathbf{Y} \right) + \Re \left(\mathbf{W} \right) \Im \left(\mathbf{Y}\right)$. We initialize the first-order weights with the conjugate transpose of the DFT matrix as it coincides with the inverse such that $\mathbf{W}_1 = \mathcal{F}^{-1} = \mathcal{F}^*$. Accordingly, the first-order computation in the first iteration of the training computes the inverse DFT transformation $\mathbf{W}_1\mathbf{Y}\mathbf{W}_1^T$ which is indeed the proxy mapping computation. Thereafter, the proxy mapping is fine-tuned jointly with the support estimator part of the network where the first operational layer takes the obtained proxy in 2D, i.e., $\widetilde{\mathbf{X}} = \phi(\mathbf{Y})$ as the input.

\begin{figure}[!hb]
    \centering
    \includegraphics[width=0.91\linewidth]{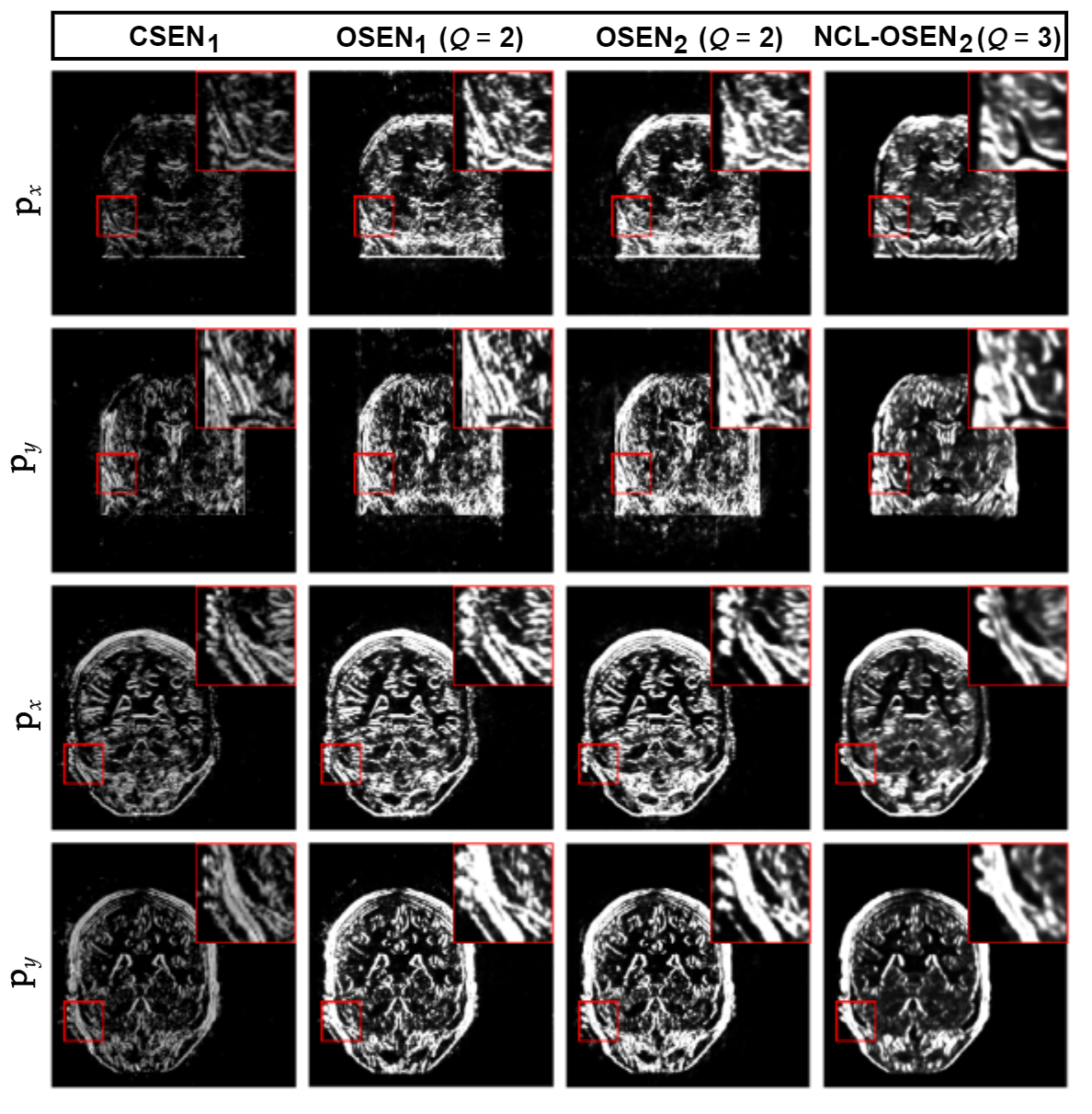}
    \caption{Estimated probability maps of support locations in the gradient domain for the $2^\text{nd}$ and $5^\text{th}$ row MRI images in Fig. \ref{fig:tv_results} when $\text{MR} = 0.25$. }
\label{fig:tv_support_results}
\end{figure}

In the experiments, we use the Diencephalon Challenge (Mid-brain) dataset \cite{mri_dataset} consisting of already partitioned 35 and 12 patients for training and testing, respectively. For both sets, we remove empty beginning and ending slices and include the slices where there are clearly visible parts of the anatomy. After the preprocessing, there are collected $6045$ training and $2048$ testing samples with the size of $256 \times 256$. From the training set, we separate another around 1000 samples for validation. Next, two-channel proxies are constructed as previously described: $\left \{ \widetilde{\mathbf{X}}_x^\text{train} =  \nabla_x \widetilde{\mathbf{S}}^\text{train}, \widetilde{\mathbf{X}}_y^\text{train} =  \nabla_y \widetilde{\mathbf{S}}^\text{train} \right \}$, where the gradient operation is applied to rough estimations obtained by the zero-padded inverse DFT which is equivalent to mapping with $\mathbf{A}^T$. Corresponding ground-truth masks are obtained by thresholding the gradient magnitudes of the original images by setting $\tau_1 = 0.04$. When proxies are fed into the OSENs, probability maps $\left (\mathbf{p}_x, \mathbf{p}_y \right )$ are obtained by the networks, which are then used in computation of the cost weights in \eqref{eq:weighted_TV_Gamma} for the proposed weighted TV reconstruction by setting $\epsilon = 0.2$.

Reconstruction performances obtained by the proposed approach and compared methods are presented in Table \ref{tab:tv_results}. It is shown that the proposed approach outperforms CSENs and the reconstruction accuracy is significantly improved. Especially for $\text{MR} = 0.1$, the improvement over the baseline approach is 2 dB in PSNR using $\text{OSEN}_2$ with $Q = 2$. The NCL module further boosts the recovery performance as any NCL configuration can achieve greater than $45$ dB when $\text{MR} = 0.25$. For low $\text{MR}$, the gap is more significant between the $\text{OSEN}_1$ and $\text{NCL-OSEN}_1$ where it can reach $0.25$ dB. It is also crucial to point out that the introduced learning-aided CS recovery approach improves the performance without bringing additional computational complexity to the SR algorithm. Because the computational complexity of performing direct mapping via OSENs is negligible compared to iterative reconstruction. For instance, TV-based (baseline) recovery has taken $412.73$, $386.71$, and $378.88$ seconds for $\text{MR} = 0.05$, $0.1$, and $0.25$, respectively, whereas $\text{OSEN}_2 (Q=3)$ has taken only 13 milliseconds. Similarly, the NCL module complexity is also negligible since the first-order computation replaces the inverse DFT transformation that is nonetheless computed in every acquisition scheme. The reconstructed MRI images are shown in Fig. \ref{fig:tv_results}. Accordingly, it is observed that learning-aided reconstructions both with CSEN and OSEN improve the overall reconstructed image quality and this shows that the output probability maps indeed can be used in the recovery of CS signals to enhance the performance of traditional model-based approach such as TV-based $\ell_1$-minimization. Moreover, investigating the zoomed regions, one can say that the details are better preserved in the proposed OSEN approaches compared to the CSEN. Recovery performances of support locations in gradient domain are presented in Table \ref{tab:mri_support_results}. It is clear that the proposed approach outperforms CSENs in the SE by more than $10\%$ and $15\%$ considering $F_1$ and $F_2$ scores, respectively. Produced output probability maps by the methods are illustrated in Fig. \ref{fig:tv_support_results}. Accordingly, estimated support maps by the OSENs have sharper details and supports at the edges are accurately estimated, whereas CSENs produce significantly higher false negatives yielding lower sensitivity levels in the SE. Moreover, the NCL-OSEN approach precisely estimates regional shapes in the probability maps as observed in Fig. \ref{fig:tv_support_results}.

\subsection{Ablation Study I: Localized versus Non-localized Kernels}

In Table \ref{tab:local_ablation}, we compare localized and non-localized kernel configurations in the proposed OSENs. Accordingly, it is shown that the usage of non-localized kernels often improves the performance of OSENs in three different applications. Especially for the $\text{OSEN}_1$ over MNIST dataset, the improvement is significant using smaller $Q$ values. It is worth mentioning that even though the highest classification accuracy is obtained using the non-local kernels in representation-based classification problem using Yale-B dataset, for $Q=1$ and $Q=3$ settings, OSENs with localized kernel configurations can obtain comparable accuracies.

\begin{table}[h!]
\caption{Performance of the proposed approach with localized (OSEN*) and non-localized (OSEN) kernels. Measurement rates are fixed to their smallest values in each application: $\text{MR}=0.01$ for Yale-B classification and $0.05$ for the others.}
\vspace{-0.2cm}
\label{tab:local_ablation}
\centering
\begin{tabular}{@{}cccc@{}}
\toprule
\multirow{2}{*}{\textbf{Method}} & \textbf{MNIST}      & \textbf{Yale-B}     & \textbf{CS MRI} \\
                                 & \boldmath$F_1$\textbf{-Score} ($\%$) & \textbf{Accuracy ($\%$)} & \textbf{PSNR (dB)} \\ \midrule
\textbf{OSEN*}$_1$ ($Q = 1$)                     & 74.82              & 36.20              & 29.81           \\
\textbf{OSEN}$_1$ ($Q = 1$)                      & 84.38              & 36.20              & 30.19           \\
\textbf{OSEN*}$_1$ ($Q = 2$)                     & 80.97              & 42.24              & 30.00           \\
\textbf{OSEN}$_1$ ($Q = 2$)                      & 87.91              & 43.02              & 30.33           \\
\textbf{OSEN*}$_1$ ($Q = 3$)                     & 82.37              & 39.56              & 30.03           \\
\textbf{OSEN}$_1$ ($Q = 3$)                      & 86.77              & 41.53              & 30.32           \\
\textbf{OSEN*}$_2$ ($Q = 1$)                     & 84.42              & 43.80              & 30.09           \\
\textbf{OSEN}$_2$ ($Q = 1$)                      & 85.85              & 41.86              & 30.38           \\
\textbf{OSEN*}$_2$ ($Q = 2$)                     & 85.97              & 42.88              & 30.18           \\
\textbf{OSEN}$_2$ ($Q = 2$)                      & 87.91              & 46.85              & 30.48           \\
\textbf{OSEN*}$_2$ ($Q = 3$)                     & 86.23              & 41.36              & 30.19           \\
\textbf{OSEN}$_2$ ($Q = 3$)                      & 88.13              & 38.92              & 30.49           \\ \bottomrule
\end{tabular}
\end{table}

\subsection{Ablation Study II: CSEN versus NCL-CSEN}

In this ablation study, we investigate the performance improvement obtained by the proposed NCL module when it is integrated into the previous support estimator. Accordingly, we fix the support estimator backbone of the framework and the NCL module is connected to the input of the CSEN. As shown in Table \ref{tab:ncl_ablation}, the overall performance is significantly enhanced by the NCL module. Incrementing the $Q$-value has further improved the performance in the majority of MR cases for different applications.

\begin{table}[h!]
\caption{Performance improvements are presented when the NCL module is integrated into the CSEN approach. Measurement rates are set to $\{0.05, 0.1, 0.25\}$ for SE over MNIST and CS-MRI; $\{0.01, 0.0.05, 0.25\}$ for Yale-B classification in (a), (b), and (c), respectively.}
\label{tab:ncl_ablation}
\centering
\setlength{\tabcolsep}{1pt} 
\renewcommand{\arraystretch}{0.8} 
\begin{tabular}{ccccc}
\toprule
& \multirow{2}{*}{\textbf{Method}} & \textbf{MNIST}      & \textbf{Yale-B}     & \textbf{CS MRI} \\
                                 & & \boldmath$F_1$\textbf{-Score} ($\%$) & \textbf{Accuracy ($\%$)} & \textbf{PSNR (dB)} \\ \midrule

\multicolumn{1}{c}{\multirow{8}{*}{\rotatebox[origin=c]{90}{(a)}}} & $\textbf{CSEN}_1$ \cite{csen} & $78.86$ & $25.69$ & $29.90$ \\
                                       
& \textbf{NCL} ($Q = 1$) \textbf{- CSEN}$_1$ & $89.52$ & $53.59$ & $30.41$ \\

& \textbf{NCL} ($Q = 2$) \textbf{- CSEN}$_1$ & $89.76$ & $54.0$ & $30.41$ \\

& \textbf{NCL} ($Q = 3$) \textbf{- CSEN}$_1$ & $89.80$ & $52.58$ & $\bm{30.43}$ \\

& $\textbf{CSEN}_2$ \cite{csen} & $83.37$ & $34.71$ & $30.05$ \\

& \textbf{NCL} ($Q = 1$) \textbf{- CSEN}$_2$ & $89.75$ & $\bm{56.14}$ & $30.40$ \\

& \textbf{NCL} ($Q = 2$) \textbf{- CSEN}$_2$ & $89.94$ & $55.93$ & $30.41$ \\

& \textbf{NCL} ($Q = 3$) \textbf{- CSEN}$_2$ & $\bm{89.99}$ & $50.34$ & $30.42$ \\ \midrule

\multicolumn{1}{c}{\multirow{8}{*}{\rotatebox[origin=c]{90}{(b)}}} & $\textbf{CSEN}_1$ \cite{csen} & $84.04$ & $87.12$ & $34.71$ \\

& \textbf{NCL} ($Q = 1$) \textbf{- CSEN}$_1$ & $93.44$ & $92.10$ & $35.13$ \\

& \textbf{NCL} ($Q = 2$) \textbf{- CSEN}$_1$ & $93.51$ & $93.46$ & $35.16$ \\

& \textbf{NCL} ($Q = 3$) \textbf{- CSEN}$_1$ & $93.51$ & $93.22$ & $\bm{35.21}$ \\

& $\textbf{CSEN}_2$ \cite{csen} & $87.47$ & $87.86$ & $34.89$ \\

& \textbf{NCL} ($Q = 1$) \textbf{- CSEN}$_2$ & $93.53$ & $92.85$ & $35.17$ \\

& \textbf{NCL} ($Q = 2$) \textbf{- CSEN}$_2$ & $93.59$ & $94.27$ & $35.19$ \\

& \textbf{NCL} ($Q = 3$) \textbf{- CSEN}$_2$ & $\bm{93.60}$ & $\bm{94.64}$ & $35.15$ \\ \midrule

\multicolumn{1}{c}{\multirow{8}{*}{\rotatebox[origin=c]{90}{(c)}}} & $\textbf{CSEN}_1$ \cite{csen} & $89.65$ & $97.22$ & $44.36$ \\

& \textbf{NCL} ($Q = 1$) \textbf{- CSEN}$_1$ & $\bm{95.83}$ & $97.56$ & $45.12$ \\

& \textbf{NCL} ($Q = 2$) \textbf{- CSEN}$_1$ & $95.79$ & $97.93$ & $45.19$ \\

& \textbf{NCL} ($Q = 3$) \textbf{- CSEN}$_1$ & $95.78$ & $\bm{98.44}$ & $\bm{45.33}$ \\

& $\textbf{CSEN}_2$ \cite{csen} & $91.41$ & $97.19$ & $44.51$ \\

& \textbf{NCL} ($Q = 1$) \textbf{- CSEN}$_2$ & $95.81$ & $97.63$ & $45.05$ \\

& \textbf{NCL} ($Q = 2$) \textbf{- CSEN}$_2$ & $95.77$ & $97.93$ & $45.08$ \\

& \textbf{NCL} ($Q = 3$) \textbf{- CSEN}$_2$ & $95.78$ & $98.41$ & $45.11$ \\ \bottomrule
\end{tabular}
\end{table}

\section{Conclusion}
\label{conclusion}

In this study, we have proposed novel support estimator approaches called OSENs. First, an OSEN can learn a direct mapping for support sets which eliminates the need for performing a prior SR task, unlike traditional methods. Next, the proposed OSEN approach consisting of operational layers with super neurons has several improvements over the traditional convolutional layers of a CSEN. The super (generative) neuron model can learn non-linear transformation functions for each kernel element and the kernel locations are jointly optimized with the self-organized transformation functions. An extended set of experiments has shown that non-localized kernels and operational layers have significantly improved the SE performance compared to their traditional convolutional layers/kernels. Moreover, thanks to the introduced learned and non-linear proxy mapping layer using Self-GOPs, it is possible to directly estimate the support set or class labels from measurement samples using end-to-end NCL-OSENs. In this way, the proposed NCL module can further enhance the performance by fine-tuning the proxy mapping layer together with the SE part of the network. 

We have evaluated the proposed approach considering three different applications: i. SE from CS measurements for the cases where the signal is sparse in the spatial domain, ii. face recognition using representation-based classification, and iii. learning aided CS MRI reconstruction. In these applications, it has been shown that the proposed approach can achieve state-of-the-art performance levels requiring minimum computational complexity. Especially in low MRs, the performance gap widens significantly. Because of the proposed direct SE scheme, it is sufficient to use compact networks in OSEN architectures with maximized learning ability thanks to the operational layers with super neurons. A crucial advantage is that compact OSENs can operate under limited training data with an elegant performance, and therefore, besides their computational efficiency compared to deep learning techniques and traditional iterative SRC approaches, they do not suffer from well-known “over-fitting” problems.

Note further the fact that the applications of the proposed OSEN approach presented in this work are not limited only to, for example, MRI reconstruction or face recognition, but its usability for different applications is possible. Similarly, the introduced Self-GOP approach in the NCL module is a generalization of MLPs and it can provide state-of-the-art performance levels in different challenging classification and regression problems thanks to the ability to learn nonlinear transformation functions. Overall, those will be our future research topics. We presented a support estimation network family, and its application areas can be considered in signal recovery within the scope of non-blind linear degradation such as compressive sensing, where the sensing matrix is fixed and known by its nature. However, in future work, we also plan to investigate the possibility of increasing the performance of sparse signal recovery via deep unfolding-based bling image restoration tasks where the linear degradation is unknown and possibly dynamic changing from measurement to measurement \cite{mou2022deep, li2023efficient}.

\section*{Acknowledgment}
This work has been supported by the Business Finland project Advanced Machine Learning for Industrial Applications (AMaLIA) under NSF IUCRC Center for Big Learning.

\ifCLASSOPTIONcaptionsoff
  \newpage
\fi



%

\bibliographystyle{IEEEtran}
\bibliography{IEEEtran}




\end{document}